%% file: main.tex
\documentclass[10pt,twocolumn,letterpaper]{article}

\usepackage[pagenumbers]{iccv} % To force page numbers, e.g. for an arXiv version


\definecolor{iccvblue}{rgb}{0.21,0.49,0.74}
\usepackage[pagebackref,breaklinks,colorlinks,allcolors=iccvblue]{hyperref}

\usepackage{graphicx}
\usepackage{amsmath}
\usepackage{amssymb}
\usepackage{booktabs}
\usepackage{enumitem}
\usepackage{multirow}
\usepackage{pifont}
\usepackage{bm}
\usepackage{comment}
\usepackage{adjustbox}
\usepackage{caption}
\usepackage{float}
\usepackage{csquotes}
\usepackage{makecell}
\usepackage[accsupp]{axessibility}  % Improves PDF readability for those with disabilities.
\newcommand\blfootnote[1]{
    \begingroup
    \renewcommand\thefootnote{}\footnote{#1}
    \addtocounter{footnote}{-1}
    \endgroup
}

\def\paperID{13391} % *** Enter the Paper ID here
\def\confName{ICCV}
\def\confYear{2025}
\def\name{VOccl3D}
\title{\name: A Video Benchmark Dataset for 3D Human Pose \\ and Shape Estimation under real Occlusions}

\author{Yash Garg, Saketh Bachu, Arindam Dutta, Rohit Lal$^{\dagger}$, Sarosij Bose,\\ Calvin-Khang Ta$^{\ddagger}$,  M. Salman Asif,  Amit Roy-Chowdhury \\
University of California, Riverside\\
{\tt\small \{ygarg002,sbach008,adutt020,rlal011,sbose007,cta003,sasif,amitrc\}@ucr.edu}
}

\begin{document}
% \maketitle
\input{sec/teaser}

\input{sec/0_abstract}    
\input{sec/1_intro}
\input{sec/2_related}

\input{sec/3_method}
\input{sec/4_experiments}
\input{sec/6_conclusion}
% \clearpage
% Appendix Section
% \appendix
% \section{Appendix}
% \input{sec/7_appendix} % If you have a separate file for appendix
{
    \small
    \bibliographystyle{ieeenat_fullname}
    \bibliography{main}
}
\clearpage
\input{suppl_under_main}

\end{document}

%% file: sec/teaser.tex
\newcommand{\teaserCaption}{
{\bf BEDLAM: }
Bodies Exhibiting Detailed Lifelike Animated Motion.
}

\twocolumn[{
    \renewcommand\twocolumn[1][]{#1}
    \maketitle
    \centering
    \vspace{-2em}
    \begin{minipage}{\textwidth}
        \centering
        \includegraphics[trim=000mm 000mm 000mm 000mm, clip=False, width=\linewidth]{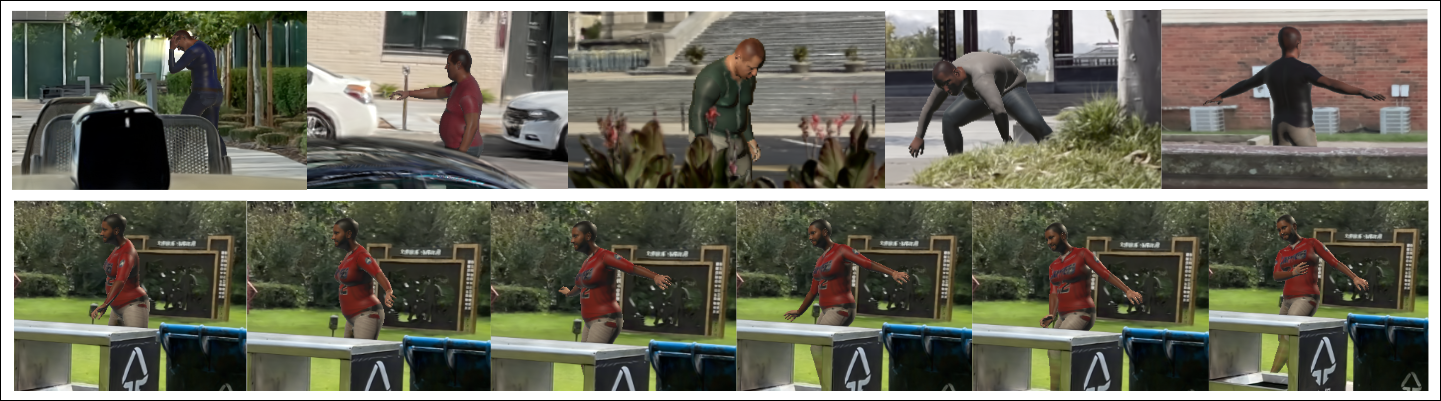}
    \end{minipage}
    \vspace{-0.5 em}
    \captionof{figure}{We propose \textbf{\name{}}, a large-scale synthetic video dataset specifically designed for training and evaluating algorithms for 3D human pose and shape estimation (HPS) in realistic occlusion scenarios. \name{} comprises over 250,000 frames, with a total runtime exceeding 2 hours and 30 minutes. Compared to previous occlusion-based datasets, \name{} has diverse body shapes, textures, and most importantly, significant and realistic occlusions within the scenes. In addition to body shape and pose, \name{} provides other necessary ground truth information, such as bounding boxes, body part segmentations, and human silhouettes. The top row illustrates various frames from our dataset, showcasing diverse occlusions, clothing textures, and motions, while the bottom row represents a sequence of frames from a video sequence within our dataset.
    % \amit{Describe a little about these images; are you trying to show discrete frames in the top row and a sequence in the bottom? Clratiy is essential; you are not sitting with the reviewer and you have a few minutes to win or lose them.}
    }
    \label{fig:teaser}
    \vspace{2.2em}
}]

%% file: sec/0_abstract.tex
\begin{abstract}
Human pose and shape (HPS) estimation methods have been extensively studied, with many demonstrating high zero-shot performance on in-the-wild images and videos. However, these methods often struggle in challenging scenarios involving complex human poses or significant occlusions. Although some studies address 3D human pose estimation under occlusion, they typically evaluate performance on datasets that lack realistic or substantial occlusions, e.g., most existing datasets introduce occlusions with random patches over the human or clipart-style overlays, which may not reflect real-world challenges. To bridge this gap in realistic occlusion datasets, we introduce a novel benchmark dataset, \name, a \textbf{V}ideo-based human \textbf{Occ}lusion dataset with \textbf{3D} body pose and shape annotations. Inspired by works such as AGORA and BEDLAM, we constructed this dataset using advanced computer graphics rendering techniques, incorporating diverse real-world occlusion scenarios, clothing textures, and human motions. Additionally, we fine-tuned recent HPS methods, CLIFF and BEDLAM-CLIFF, on our dataset, demonstrating significant qualitative and quantitative improvements across multiple public datasets, as well as on the test split of our dataset, while comparing its performance with other state-of-the-art methods. Furthermore, we leveraged our dataset to enhance human detection performance under occlusion by fine-tuning an existing object detector, YOLO11, thus leading to a robust end-to-end HPS estimation system under occlusions. Overall, this dataset serves as a valuable resource for future research aimed at benchmarking methods designed to handle occlusions, offering a more realistic alternative to existing occlusion datasets. See the Project page for code and dataset:\url{https://yashgarg98.github.io/VOccl3D-dataset/}
% Our dataset also provides annotations for other modalities under occlusion, including segmentation, body-part segmentation, human bounding boxes, and 2D body keypoints.
% \vspace{-4ex}
\end{abstract}
\blfootnote{$\dagger$ Currently at NASA MSFC IMPACT. $\ddagger$ Currently at Dolby Laboratories. Work done while the authors were at UCR.}

%% file: sec/1_intro.tex
\section{Introduction}
\label{sec:intro}

%  1st para - brief about HPS methods
Monocular 3D human pose and shape (HPS) estimation is a complex yet essential task in computer vision. It has applications in surveillance \cite{surveillance1, surveillance2}, autonomous robotics \cite{pose_aut1, pose_aut2, pose_aut3}, human motion analysis \cite{gait_1, gait_2}, and clinical assessment \cite{clinical_1, clinical_2, koleini2025biopose}. Since the introduction of the neural network-based Human Mesh Recovery Network (HMR) \cite{Kanazawa2017EndtoEndRO}, numerous approaches have advanced the field by enabling accurate estimation of SMPL\cite{SMPL} pose and shape from a single RGB image. Recent methods \cite{Arnab_CVPR_2019, li2022cliff, meva, kocabas2019vibe, wham, stride} demonstrate impressive zero-shot HPS performance on in-the-wild RGB images and video sequences. However, these methods still face limitations in achieving high performance in challenging scenarios, such as complex human poses and under significant occlusions. 
% With the advent of deep learning and the availability of large-scale datasets, human pose and shape estimation (HPSE) has achieved significant advancements in both 2D and 3D. Modern deep learning models can now estimate human key points and body shape with impressive accuracy and efficiency. 

% One of the recent milestones that propelled this field further is the introduction of the Human Mesh Recovery Network (HMR) \cite{Kanazawa2017EndtoEndRO} which is a convolutional neural network (CNN) that estimates both SMPL pose and shape of the human, given a single RGB image as the input. After this, there was a heavy influx of new methods which performed HPSE in several interesting settings like single-person vs multi-person, single-view vs multi-view, images vs videos, top-down vs bottom-up, indoor vs outdoor environments, etc. \sak{Cite relevant works}. 

%  2nd para - regarding occlusions in HPS and other methods
Achieving robust HPS under occlusion remains a challenging problem due to the contextual ambiguity of the occluded body parts.
% These occlusions generally fall under two categories: human-to-human and human-to-object occlusions.
% \amit{I do not think we have done human-to-human, so why bring this up and put up a target for the reviewers? If we have, then is ok, but add some examples} \yg{we have not handled human-to-human occlusion, I've removed the sentence describing it}
Several methods \cite{rempe2021humor, 3dnbf, Kocabas_PARE_2021, glamr} have been proposed to address the challenges caused by occlusions. Methods like PARE \cite{Kocabas_PARE_2021} focus on using visible body parts to infer occluded areas, enhancing estimation accuracy with partial views. Temporal models like HuMoR \cite{rempe2021humor} and GLAMR \cite{glamr} use generative frameworks to ensure pose continuity over time. HuMoR predicts pose distributions, while GLAMR incorporates global trajectory data to fill in missing poses. Recent methods, like STRIDE \cite{stride}, leverage a large-scale pre-trained motion prior model to achieve temporally coherent pose reconstruction under occlusions. While these methods effectively leverage visible information and temporal continuity, they still struggle under severe occlusions due to limited exposure to occluded scenarios in their training data.

\begin{figure}[t]
    \centering
    \includegraphics[width=\linewidth]{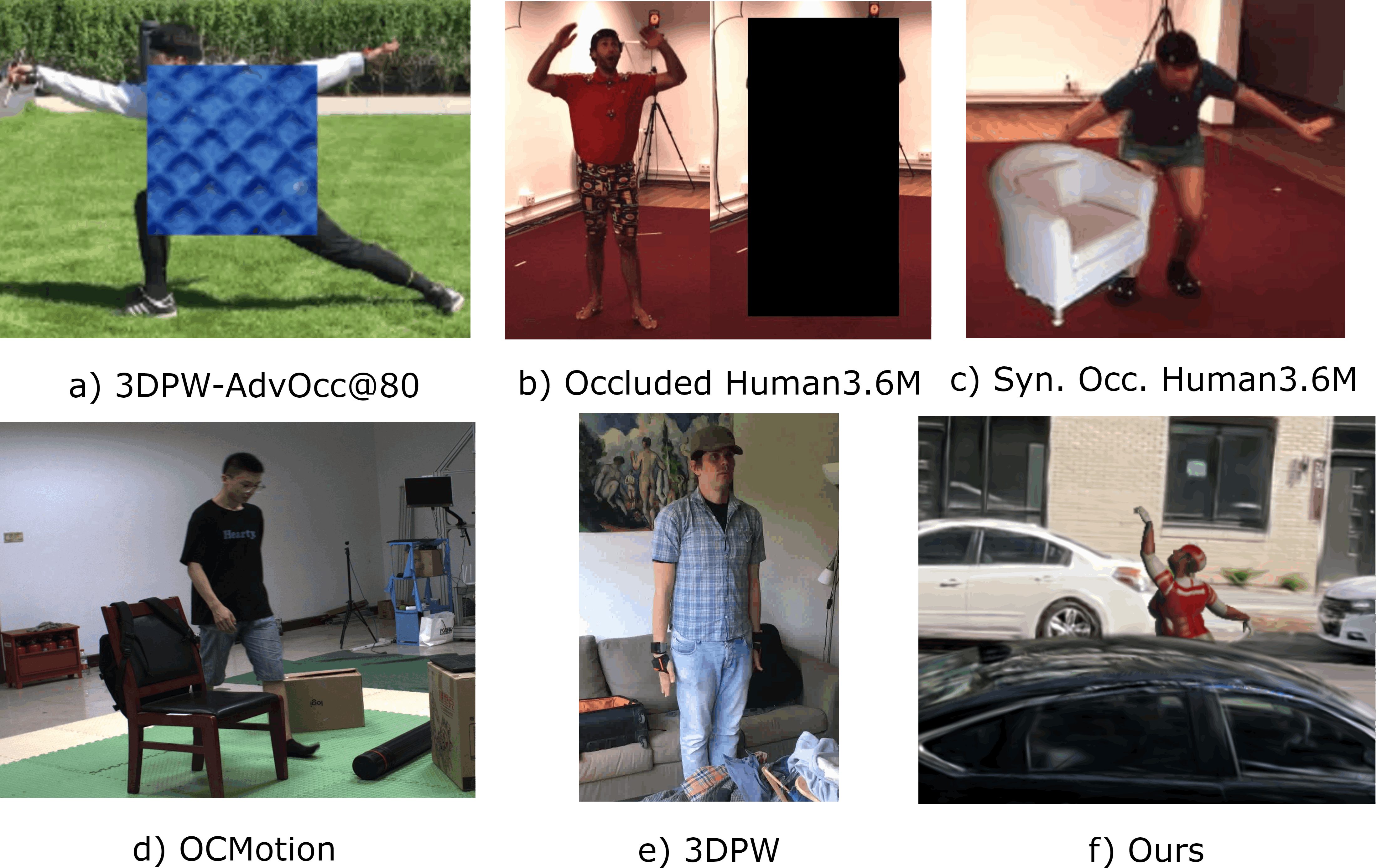}
    \caption{Samples from various occlusion-based HPS datasets. The \textbf{top row} displays datasets with artificial, patch-based occlusions applied to existing datasets, including 3DPW-AdvOcc \cite{3dpw, 3dnbf}, Occluded Human3.6M \cite{h36m, stride}, and Synthetic Occlusion Human3.6M \cite{h36m, 3doh50k}. The \textbf{bottom row} presents samples from two datasets with natural occlusions: OCMotion \cite{ocmotion} and 3DPW \cite{3dpw}. Notably, the top-row images exhibit unrealistic occlusions, which lack the realism of naturally occurring occlusions. The bottom row images contain natural occlusions but are sparse and infrequent. Compared to existing datasets, our proposed \name{} dataset features more realistic occlusions.}
    \vspace{-2ex}
    \label{fig:samples_datasets}
\end{figure}

%  3rd para - what is missing in current literature and what are we proposing
Evaluations of the existing methods are often limited to occluded datasets that lack scene diversity \cite{i3db}, involve moderate or lower levels of occlusion \cite{ocmotion, 3dpw}, or use patch-based occlusion datasets for training and inference \cite{3dnbf}, as illustrated in Figure~\ref{fig:samples_datasets}. Most importantly, there remains an urgent need for a realistic, diverse, and significantly occluded human pose and shape dataset to advance the task of 3D HPS estimation.
%  This is backed by the fact that most of the widely used public video datasets like 3DPW, OCMotion, Human 3.6M, etc do not contain severe occlusions in their samples which naturally limits the capability of any models trained on them to handle occlusions. Thus, to sustain this fast-moving field of HPSE, datasets containing occlusion at a large scale are the need of the hour.
To address this gap, our work proposes~\name{}, a large-scale video dataset with synthetic humans in real scenes for 3D human pose and shape estimation.
% We demonstrate both qualitatively and quantitatively that recent HPS estimation methods perform poorly on our dataset. 
Following prior works such as BEDLAM \cite{black2023bedlam} and Synthmocap \cite{hewitt2024look}, we use an advanced computer graphics engine to render a highly realistic synthetic dataset. Prior research has shown that fine-tuning models on synthetic datasets can significantly improve HPS estimation performance on real-world datasets \cite{black2023bedlam}. Synthetic datasets provide \textquote{perfect} ground-truth annotations and eliminate the need for costly sensors, making it computationally efficient to generate large-scale datasets.

% 4th para - how are we creating dataset and its attributes - also what's crucial difference while rendering from others
Our proposed dataset includes approximately 250,000 images and 400 video sequences. To introduce diversity, we incorporate human motion samples from the AMASS \cite{AMASS} dataset, over 200 distinct clothing textures for sequences containing both male and female genders, and 40 real background scenes with occlusions. Additionally, we provide occlusion labels for each body joint. Unlike prior methods \cite{black2023bedlam, patel2021agora} which render synthetic environments, we utilize 3D Gaussian splatting (3DGS) \cite{3dgs} to create 3D representations of background scenes. This approach enhances the realism of rendered scenes, making them closely resemble real-world captured data. Reconstructing background scenes using 3DGS offers the flexibility to create domain-specific datasets, such as for agriculture, street or indoor environments using raw RGB video captures without relying on costly or limited graphic assets. Figure~\ref{fig:teaser} showcases samples from our dataset, illustrating the diversity in clothing, environments, human motions, and occlusions.
% We render the synthetic image sequences in Unity Engine at 30 frames per second (fps) with various camera focal lengths.

% 5th para - experimentation and what we showed
We fine-tuned the CLIFF~\cite{li2022cliff} and BEDLAM-CLIFF~\cite{black2023bedlam} models on our dataset to demonstrate that training with synthetic data enhances the performance of existing HPS estimation methods under occlusion. We evaluated state-of-the-art HPS estimation methods on the test split of our dataset across three occlusion levels: high, medium, and low. The results show that existing methods perform poorly under high occlusion, whereas our fine-tuned models achieve significant improvements. To assess performance on real-world data, we perform evaluations on the 3DPW and OCMotion datasets, which contain low-level occlusions. Additionally, we created a variant of the 3DPW dataset by adding black patches to evaluate the robustness against high-occlusion real scenarios. Our fine-tuned models outperformed recent HPS methods on these real-world datasets. Furthermore, our experiments highlight the importance of an effective human object detector for improving HPS performance under occlusion. To address this, we fine-tuned the YOLO11\cite{yolo11_ultralytics} object detector using our dataset, enhancing its performance under occluded scenarios. \name{} is available for researchers to benchmark and evaluate occlusion-aware methods.

% Furthermore,~\name{} provides annotations for various attributes under occlusion, including human silhouettes, body-part segmentations, 2D keypoints, and bounding boxes for human detection. This dataset serves as a valuable resource for evaluating methods designed to perform under occlusion, providing a more realistic alternative to existing datasets. 
In summary, we make the following key contributions:
\begin{enumerate}[topsep=2pt]
    \item We propose \textbf{\name{}}, a novel large-scale, realistic video-based dataset of occluded synthetic humans in real scenes for 3D human pose and shape estimation.
    \item We demonstrate both qualitatively and quantitatively that existing HPS estimation methods fine-tuned with~\name{} outperform other methods on real-world datasets with occlusions.
    \item We improved the performance of the human object detector under occlusion scenarios by leveraging the~\name{} dataset which is a crucial component for a robust HPS estimation. 
    \item \name{} can serve as a benchmark dataset for evaluating methods specifically designed to perform under occlusion across various tasks, including human and body-part segmentation, 2D/3D pose estimation, and human bounding box detection.
\end{enumerate}
% \amit{The entire intro should not exceed 2 pages - compress it down.}

%% file: sec/2_related.tex
 \section{Related Works}
\label{sec:formatting}
% \amit{This is very long. Keep it to 1-1.5 cols. You do not need to describe every paper.}

\noindent\textbf{Synthetic Data Generation for Human Pose Estimation.}
Human pose estimation is crucial in computer vision, but state-of-the-art methods depend on costly, labor-intensive labeled datasets. Several notable synthetic datasets have advanced research in human pose estimation. SynBody \cite{yang2023synbody} provides a large-scale synthetic dataset for human mesh recovery and view synthesis, while RePoGen \cite{purkrabek2023improving} enables fine-grained control over pose and viewpoint to generate rare, complex poses. Human3.6m~\cite{h36m} provides additional small mixed reality dataset by inserting 3D animation models with background scenes. BEDLAM \cite{black2023bedlam} further highlights that models trained solely on synthetic data can outperform real-data-trained counterparts, emphasizing the importance of high-quality synthetic datasets for transferable models. Further, PressurePose \cite{clever2020bodies} simulates interactions between articulated and soft bodies, capturing fine-grained contact dynamics. SynthMocap \cite{hewitt2024look} extends this by providing expressive synthetic data with detailed body, hand, and face movements. Despite their success, these synthetic datasets lack emphasis on significant occlusions, a key real-world challenge. Addressing this limitation, our work proposes a synthetic dataset tailored to 3D pose estimation under heavy and realistic occlusion, aiming to bridge this gap and enhance robustness in occlusion-prone environments.

\noindent\textbf{Image and Video based Human Pose Estimation.}
Estimating 2D/3D pose and shape from single RGB image has widely been explored in previous research \cite{Sharma_2019_ICCV, li2022cliff, WehRud2021, pcls, black2023bedlam}. Methods such as \cite{Sharma_2019_ICCV} use a conditional variational autoencoder for 2D-to-3D lifting in pose estimation. The pose estimation approach proposed in \cite{WehRud2021} applies normalizing flows \cite{norm1} for 2D-3D mapping. Large-scale datasets like BEDLAM \cite{black2023bedlam} have improved pose estimation models like CLIFF \cite{li2022cliff} and HMRNet \cite{Kanazawa2017EndtoEndRO}. However, these models struggle with generalizing to unseen scenarios with severe occlusions due to limited training on such cases.
% \noindent\textbf{Image-based Human Pose Estimation} Estimating 2D/3D pose and shape from single RGB images has widely been explored in previous research \cite{Sharma_2019_ICCV, li2022cliff, WehRud2021, pcls, black2023bedlam}. Pioneering works, such as \cite{Sharma_2019_ICCV}, utilize a conditional variational autoencoder to perform 2D-to-3D lifting for 3D pose estimation. \cite{WehRud2021} employs normalizing flows \cite{norm1} to solve the 2D-3D mapping and then utilize the inverse mapping to estimate the 3D pose. Further, the method proposed in \cite{pcls} utilizes perspective crop layers (PCLs) to improve 3D HPS by paying attention to perspective effects. Newer works like \cite{li2022cliff} proposed a method that incorporates location information from the entire frame to enhance 3D pose estimation process. With the introduction of large-scale datasets like BEDLAM \cite{black2023bedlam}, models like cliff \cite{li2022cliff} and HMRNet \cite{Kanazawa2017EndtoEndRO} have seen a tremendous improvement in performance as a result of pre-training on such datasets. However, since these models are not extensively trained for handling occlusions, they do not generalize well to unseen scenarios consisting of severe occlusions.

In addition, video-based human pose estimation has significantly improved challenging datasets. Early works like \cite{Zhou2015SparsenessMD} use the EM algorithm to estimate the 3D pose from monocular video through 2D joint uncertainty maps. The method in \cite{Pavllo20183DHP} employs dilated temporal convolutions and semi-supervised learning for 3D pose estimation, while \cite{Arnab_CVPR_2019} uses the SMPL model \cite{SMPL} to extract pose and shape parameters, refining models like HMR with bundle adjustment. VIBE \cite{kocabas2019vibe} applies adversarial learning with AMASS \cite{AMASS} for 3D pose extraction, while MEVA \cite{meva} improves accuracy and smoothness using a variational autoencoder to address VIBE’s high acceleration error. Owing to these drawbacks, HuMoR employs a conditional variational autoencoder for robust pose estimation, while the state-of-the-art method WHAM \cite{wham} integrates 2D-3D lifting and SLAM for accurate global motion estimation. While these methods perform well on various datasets, studies like \cite{stride} highlight performance drops under significant occlusions, as most datasets contain only sparse occlusions, making models struggle with unseen heavy or prolonged occlusions.

\noindent\textbf{Human Pose Estimation under Occlusion.}
Despite significant progress in HPS estimation, handling occlusions remains a major challenge. This is because, in 3D pose estimation, missing depth cues make reconstruction far more ambiguous than in 2D. Early works like \cite{occ-aware} used data augmentation with occlusion labels to enhance robustness in pose estimation. Latest methods like GLAMR \cite{glamr} use a deep generative motion infiller to handle missing poses under severe occlusions and a global optimization framework to refine motion trajectories. Additionally, some methods \cite{smoothnet, meva, attn-seq} approach the issue of missing poses as a pose refinement problem, leveraging temporal smoothness to address it effectively. SmoothNet \cite{smoothnet} introduced a temporal motion refinement network for refining the poses obtained from the image-based pose estimators to alleviate jitters. These methods perform well under sparse, infrequent occlusions across frames but struggle with natural, prolonged occlusions, as they lack training for such instances. All these works evaluate their methods on datasets with sparse occlusions, as no existing datasets contain significant occlusions. To address this gap, we introduce \name{}, a novel dataset designed specifically for 3D HPS estimation under significant occlusions. 
%Additional related works are mentioned in the supplementary.

% \begin{itemize}
%     \item existing dataset generation papers (diffusion-based methods vs rendering-based methods, real vs synthetic dataset).

%     \item existing pose-estimation papers and their datasets.

%     \item existing occlusion-handling pose estimation methods.
% \end{itemize}

%% file: sec/3_method.tex
\section{\name{} Dataset}
In this section, we outline the key components required to construct the \name{} dataset. Section~\ref{dataset:asset} details the creation of individual assets, including background scenes, human motions, and texture maps. Sections~\ref{method:render} describe the rendering process and attributes of the dataset.

\begin{figure*}[!htbp]
    \centering
    \includegraphics[width=\textwidth]{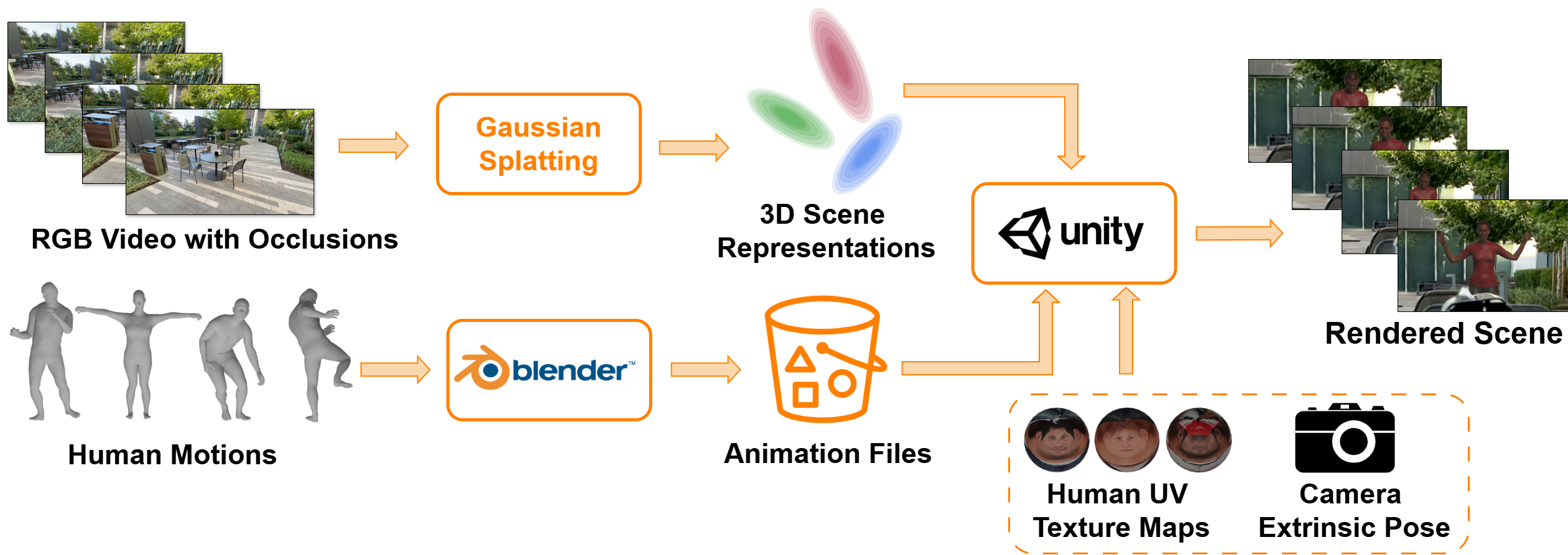}
    \caption{\textbf{An overview of our proposed dataset creation pipeline.} We generate 3D representations of natural scenes with real occlusions using 3D Gaussian Splatting \cite{3dgs}. Human motion sequences from the AMASS MoCap dataset \cite{AMASS} are processed in Blender to generate baked animation files. The generated scene representations, animation files, human texture maps, and camera extrinsic parameters are imported into the Unity rendering engine to generate video sequences of human motion under occlusion.}
    \vspace{-2ex}
    \label{fig:main_fig}
\end{figure*}

\subsection{Dataset Assets}
\label{dataset:asset}
% \subsection{Background scenes}
% \label{method:bkg}
\noindent\textbf{Background Scenes.}
In recent years, neural networks have significantly advanced 3D scene representation, enabling novel-view image synthesis. Notable approaches include Neural Radiance Fields (NeRF)~\cite{mildenhall2021nerf}, which learn a joint representation of geometry and appearance for novel-view synthesis. Another method, 3D Gaussian Splatting (3DGS)~\cite{3dgs}, represents a scene using 3D Gaussians and adopts a differentiable rasterizer for real-time rendering. These methods use multiview images to learn 3D representations, avoiding extensive capture setups. To develop our dataset, we use the 3DGS method to learn the 3D representation of the background scene.
% to  the The 3DGS methods tend to achieve higher training and rendering speed than the traditional NERF method. 
%

% [explain the 3dGS using equation]
\noindent \textbf{Preliminaries: 3D Gaussian Splatting}
% We represents a 3D scene by arranging 3D Gaussians. 
%
We represent the $i$-th Gaussian in 3DGS as: 
\begin{equation}
    G(\mathbf{p}) = o_i \, e^{-\frac{1}{2} (\mathbf{p} - \bm\mu_i)^T \bm\Sigma_{i}^{-1} (\mathbf{p} - \bm\mu_i)},
    \label{eq:gauss3d}
\end{equation}
where $\mathbf{p} \, {\in} \, \mathbb{R}^3$ is a xyz location, $o_i \, {\in} \, [0, 1]$ is the opacity modeling the ratio of radiance the Gaussian absorbs, $\bm\mu_i \, {\in} \, \mathbb{R}^3$ is the center/mean of the Gaussian, and the covariance matrix $\bm\Sigma_{i}$ is parameterized by the scale $\mathbf{S}_{i} \, {\in} \, \mathbb{R_+}^3$ along each of the three Gaussian axes and the rotation $\mathbf{R}_{i} \, {\in} \, SO(3)$ with $\bm\Sigma_{i} = \mathbf{R}_{i} \mathbf{S}_{i} \mathbf{S}_{i}^\top \mathbf{R}_{i}^\top$.
Each Gaussian is also paired with spherical harmonics~\cite{ramamoorthi2001efficient} to model the radiance emitted towards various directions.
During rendering, the 3D Gaussians are projected onto the image plane and form 2D Gaussians~\cite{zwicker2001surface} with the covariance matrix $\bm\Sigma_{i}^{\textrm{2D}} = \bm{J} \bm{W} \bm\Sigma_{i} \bm{W}^\top \bm{J}^\top$,
where $\bm{J}$ is the Jacobian of the affine approximation of the projective transformation and $\bm{W}$ is the viewing transformation. 
The color of a pixel is calculated via alpha blending the $N$ Gaussians contributing to a given pixel: 
\begin{equation}
    C = \sum_{j = 1}^{N} c_j \alpha_{j} \prod_{k=1}^{j-1} (1 - \alpha_{k}),
\end{equation}
where the Gaussians are sorted from close to far, $c_j$ is the color obtained by evaluating the spherical harmonics given viewing transform $W$, and $\alpha_j$ is calculated from the 2D Gaussian formulation (with the covariance $\bm\Sigma_{j}^{2D}$) multiplied by its opacity $o_j$.

We collected RGB videos from the large-scale open-source DL3DV dataset~\cite{dl3dv}, which contains over 10,000 videos across diverse domains, including natural and outdoor settings, educational institutions, shopping complexes, parks, hubs, cafes, and restaurants. We selected videos based on the presence of natural occlusions, such as garlands, chairs, benches, statues, cars, and bins. Our dataset captures 3D representations of approximately 40 distinct scenes, each incorporating real-world occlusions. This approach to learning 3D representation closely resembles real-world data compared to conventional methods that rely on rendering from 3D graphic asset scenes.

% \subsection{SMPLX body/Human animation}
% \label{method:human_body}
\noindent\textbf{SMPLX body/Human Animations.}
We represent the human body using the SMPL-X~\cite{pavlakos2019expressive} 3D human model, defined by the function $\mathcal{M}(\theta, \beta, \psi)$, where $\theta$ represents the pose parameters, $\beta$ the shape parameters, and $\psi$ the facial expression parameters. This function outputs a body mesh $\mathcal{M} \in \mathbb{R}^{10475 \times 3}$ with 10,475 vertices. We sample approximately 400 SMPL-X 3D human motion models from the AMASS mocap dataset~\cite{AMASS}, which contains over 11,000 motion sequences with diverse body shapes and poses. Following previous work~\cite{weng2024diffusion}, we use the pre-trained human pose prior model VPoser~\cite{pavlakos2019expressive} to assess pose difficulty.
VPoser is a Variational Auto-Encoder model that evaluates a pose $\theta$ to be challenging if its embeddings $\epsilon_{\theta}$ have larger norm, i.e. $||\epsilon_{\theta}||_{2}>\tau$, where $\tau$ is empirically set to 40. 
We classify a challenging pose using VPoser to avoid simple movements such as walking, standing, jogging, etc. To prepare human animations for rendering, we use Blender graphics software to efficiently bake complex geometric animations over time into the Alembic format for subsequent use in rendering engines.

We ensured that motion sequences contained at least 180 frames to maintain a minimum animation duration of six seconds at 30 fps. For sequences exceeding 400 frames, we discarded the first 100 frames to remove static poses at the start. To ensure that human motion remains consistently within the occluded scene, we implemented boundary constraints that stop the motion if it moves beyond these limits. These conditions prevent humans from moving too far from the occlusion area. Additionally, we applied random rotations and translations to each sequence. We store the applied transformations to further calculate the effective global orientation of the human body with respect to the world coordinates.

\begin{figure}[!ht]
    \centering
    \includegraphics[width=0.8\linewidth]{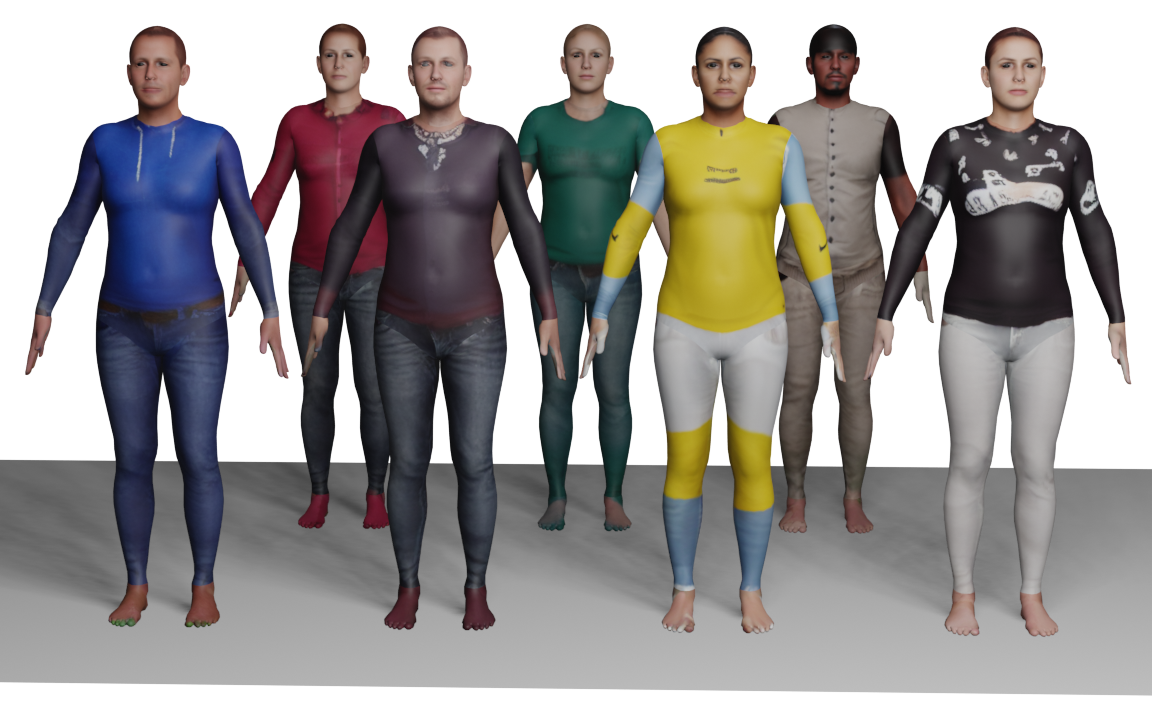}
    \caption{\textbf{Diversity of human texture in \name{}.} We use approximately 200 human texture maps from the SMPLitex dataset \cite{casas2023smplitex}, encompassing diverse clothing styles, genders, skin tones, and ethnicities.}
    \label{fig:texture-diversity}
    \vspace{-2ex}
\end{figure}

% \subsection{Human texture}
% \label{method:texture}
\noindent\textbf{Human Textures.}
We sourced human body texture scans from the open-source dataset provided by the SMPLitex method \cite{casas2023smplitex}, which estimates and manipulates the full 3D appearance of humans captured from a single image. The SMPLitex dataset provides a diverse range of human texture scans, covering various skin tones, clothing styles, and genders. To ensure sample diversity in \name{}, we select approximately 200 distinct texture scans. Figure~\ref{fig:texture-diversity} illustrates the different clothing textures applied to the SMPL-X human model in our dataset.

% \amit{why is Fig 4 before Fig 3?}

% \begin{figure*}[!ht]
%     \centering
%     \includegraphics[height=5cm, width=\textwidth]{images/modalities.png}
%     \caption{Modalities -- Todo.}
%     \label{fig:other_modalities}
% \end{figure*}

\subsection{Dataset Rendering and Attributes}
% \subsection{Rendering}
\label{method:render}

\noindent\textbf{Rendering}
As shown in Figure~\ref{fig:main_fig}, we used Unity Engine to render synthetic scenes by integrating background Gaussian splats, human animations, and body texture assets. We rendered video sequences at 30 fps with a resolution of 720×720, varying the camera field of view between 25 and 50 across different scenes. For each sequence, we stored ground-truth camera extrinsic and intrinsic parameters and saved the rendered RGB images in lossless PNG format. We positioned the camera to capture occluded views within the scene and applied motion constraints to human animation assets to maintain consistent occlusion throughout each sequence. To achieve realistic lighting, we employed Unity’s built-in DirectionalLight asset to simulate natural illumination on the human mesh.
% For an additional experiment, we have rendered a small split of multi-view datasets. We adjust the camera position in the same scene to capture human motions in an unoccluded scenario.

% \subsection{Dataset attributes}
% \label{method:dataset}
\noindent\textbf{Dataset Attributes.}
Our proposed dataset, \name{}, includes over 250,000 images and 400 video sequences with a total runtime exceeding 2 hours and 30 minutes. The dataset features 40 background scenes with various occlusions, such as cars, garlands, benches, chairs, bins, and trees. Each scene contains 10 video sequences with variations in human motions and clothing textures. The rendered sequence exhibit diversity in body shapes, skin tones, and camera poses. We provide ground-truth annotations, including camera extrinsic and intrinsic parameters, pose, shape, global orientation, translation, gender, 2D keypoints, and a binary occlusion label for each keypoint.

\noindent\textbf{Different Modalities.}
In addition to 3D pose and shape annotations, \name{} provides annotations for multiple modalities, including human silhouettes, body-part segmentation, 2D keypoint estimation, and human bounding box detection under occlusion. Researchers can use our dataset to train and evaluate methods designed to handle occlusion across these modalities.
\vspace{-2ex}
% \amit{This may be the place to cite the qualitative examples figures.}

%% file: sec/4_experiments.tex
\begin{table*}[t]
    \centering
    \resizebox{1\linewidth}{!}{
    \begin{tabular}{lccc|ccc|ccc} %{llllllllll}
    \toprule
    \multirow{2}{*}{Method}                 & \multicolumn{3}{c}{Hard-Occlusion}         & \multicolumn{3}{c}{Medium-Occlusion}     & \multicolumn{3}{c}{Low-Occlusion}       \\
                                      & MPJPE           & PA-MPJPE       & PVE             & MPJPE          & PA-MPJPE       & PVE             & MPJPE          & PA-MPJPE       & PVE            \\
    \midrule
    CLIFF~\cite{li2022cliff}           & 192.22          & 114.35         & 247.41          & 121.70         & 78.56          & 158.46          & 98.82          & 67.64          & 126.92         \\
    BEDLAM-HMR~\cite{black2023bedlam}               & 167.35          & 102.92         & 214.39          & 102.55         & 68.13          & 134.59          & 86.55          & 54.48          & 110.15         \\
    BEDLAM-CLIFF~\cite{black2023bedlam}             & 154.86          & 99.53          & 199.95          & 90.97          & 65.03          & 119.63          & 74.95          & 52.65          & 96.60          \\
    HMR2.0~\cite{goel2023humans}                   & 169.71          & 100.49         & 215.17          & 113.88         & 71.78          & 145.62          & 88.53          & 59.08          & 114.39         \\
    STRIDE with BEDLAM-CLIFF~\cite{stride}    & 155.64          & 100.44         & -               & 91.14          & 65.38          & -               & 75.02          & 53.21          & -              \\
    WHAM~\cite{wham}                     & 152.15          & 102.14         & 177.07          & 110.97         & 76.81          & 127.45          & 93.90          & 66.68          & 106.51         \\
    \midrule
    VOccl3D-B-CLIFF          & \textbf{136.34} & \textbf{89.94} & \textbf{175.92} & 82.48          & \textbf{58.78} & \textbf{106.84} & \textbf{69.46} & \textbf{46.32} & \textbf{88.19} \\
    STRIDE with VOccl3D-B-CLIFF & 136.43          & 90.28          & -               & \textbf{82.37} & 58.98          & -               & 69.65          & 46.86          & -        \\     
    \bottomrule
    
    \end{tabular}}
    \caption{\textbf{3D HPS estimation results on the test-split of \name{}.} The results show that \name{}-B-CLIFF and STRIDE, when using pseudo-labels from \name{}-B-CLIFF, significantly outperform other image- and video-based HPS estimation methods across hard, medium, and low occlusion categories. As expected, all HPS estimation methods exhibit a decline in performance as occlusion severity increases from low to medium to high. For evaluation, we use ground-truth bounding boxes. The best results are in \textbf{bold}.
}
    \label{tab:voccl3d-infer-gt}
\end{table*}

\begin{figure}[!ht]
    \centering
    \includegraphics[width=\linewidth]{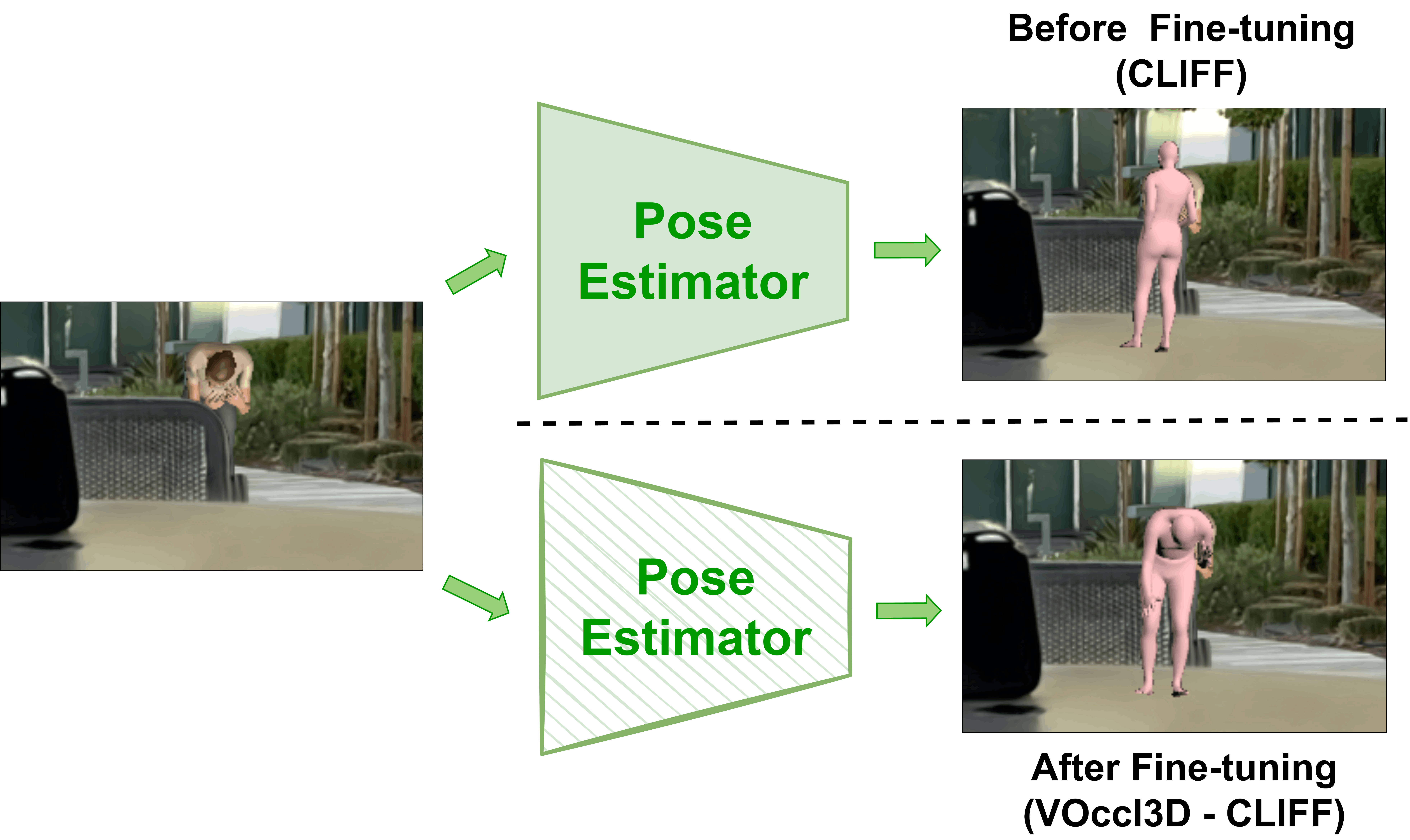}
    \caption{\textbf{Fine-tuning an off-the-shelf pose estimator.} We fine-tune a CLIFF \cite{li2022cliff} pose estimation model using our proposed \name{} dataset. We observe a significant improvement in the estimated mesh when using a fine-tuned CLIFF model using our dataset. Note: The pose estimator can be any off-the-shelf pose estimator, here we show results with the CLIFF model \cite{li2022cliff}.}
    \vspace{-3ex}
    \label{fig:yash_train_fig}
\end{figure}

\begin{table*}[]
    \centering
    \resizebox{0.9\linewidth}{!}{
    \begin{tabular}{lccc|ccc|ccc}
    \toprule
    \multirow{2}{*}{Method} & \multicolumn{3}{c}{3DPW}                         & \multicolumn{3}{c}{OcclType1-3DPW}                & \multicolumn{3}{c}{OcclType2-3DPW}                \\
                            & MPJPE          & PA-MPJPE        & PVE            & MPJPE          & PA-MPJPE        & PVE             & MPJPE          & PA-MPJPE        & PVE             \\
    \midrule
    % CLIFF*                  & 68.8           & 43.6           & 82.1           &                &                &                 &                &                &                 \\
    CLIFF~\cite{li2022cliff}                   & 73.9           & 46.4           & 87.6           & 98.15          & 62.27          & 118.47          & 99.49          & 62.16          & 119.82          \\
    BEDLAM-HMR~\cite{black2023bedlam}        & 79.0           & 47.6           & 93.1           & 108.62         & 66.53          & 128.05          & 106.19         & 64.05          & 125.66          \\
    BEDLAM-CLIFF~\cite{black2023bedlam}      & 72.0           & 46.6           & 85.0           & 98.71          & 64.26          & 117.41          & 96.80          & 61.32          & 115.33          \\
    HMR2.0~\cite{goel2023humans}            & 81.2           & 54.3           & 143.7          & 103.40         & 69.66          & 164.55          &  99.01         & 66.17          & 158.79          \\
    \midrule
    VOccl3D-B-CLIFF         & 72.0           & 47.3           & 84.5           & 95.89          & 63.43          & \textbf{114.28} & 94.36          & 60.44          & \textbf{112.01} \\
    VOccl3D-CLIFF           & \textbf{71.10} & \textbf{45.98} & \textbf{84.25} & \textbf{95.17} & \textbf{61.83} & 114.95          & \textbf{93.74} & \textbf{59.66} & 112.59 \\
     % VOccl3D-CLIFF*          & 67.4           & 43.2           & 79.8           &                &                &                 &                &                &                 \\
    \bottomrule
    \end{tabular}}
     \caption{\textbf{3D HPS estimation results on 3DPW, OcclType1-3DPW, and OcclType2-3DPW}. Since 3DPW is a real-world dataset with minimal occlusions, \name{}-CLIFF outperforms all other methods but shows only a marginal improvement over BEDLAM-CLIFF. However, on OcclType1-3DPW and OcclType2-3DPW, which contain significant occlusions on real-world dataset, our method demonstrates a notable improvement over existing approaches. The best results are noted in \textbf{bold}.}
    \label{tab:3dpw-infer-gt}
\end{table*}

\begin{figure*}[!ht]
    \centering
    \includegraphics[width=\textwidth]{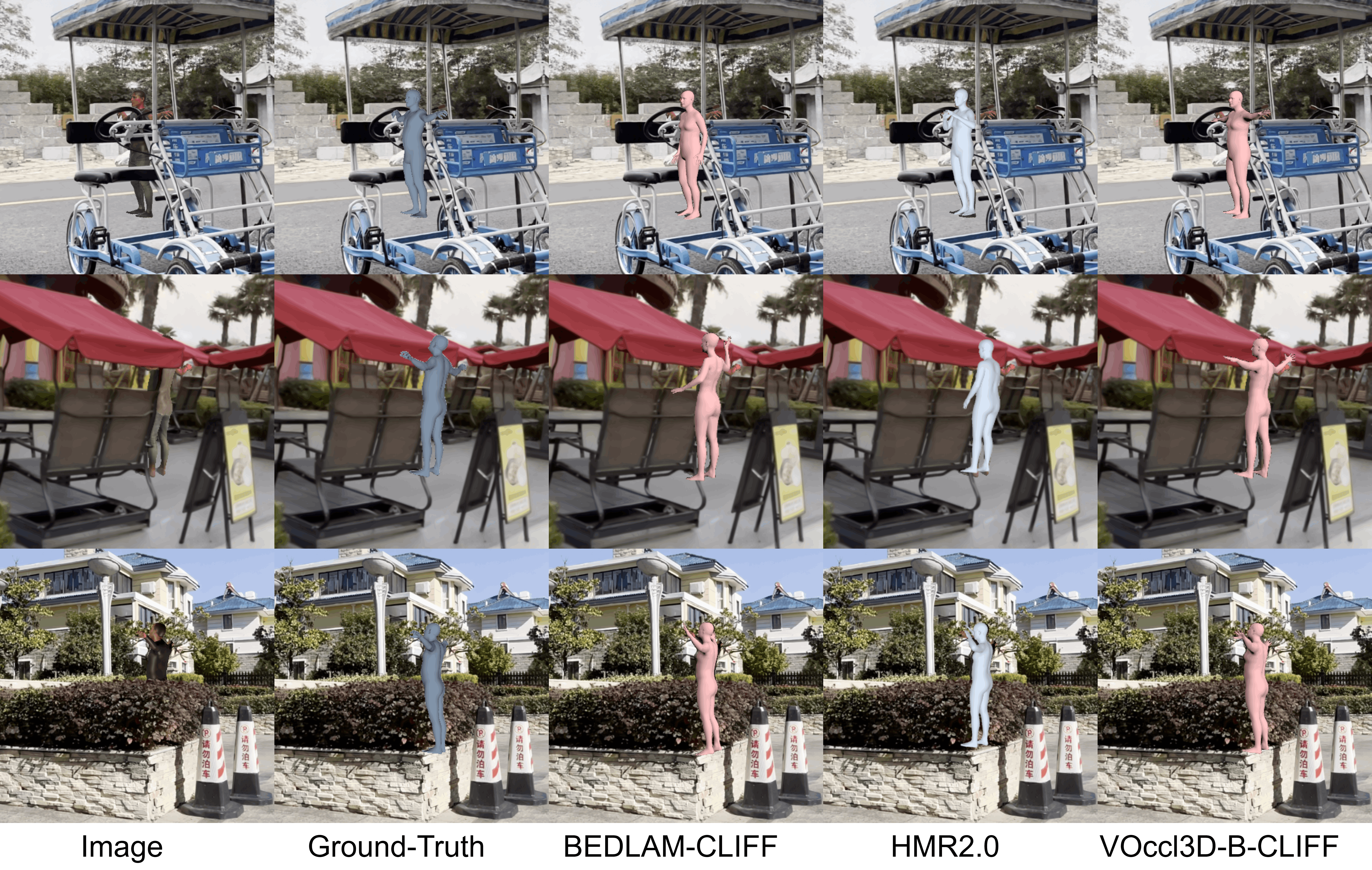}
    \caption{\textbf{Qualitative comparison of HPS estimation methods on \name{} dataset.} The first and second column shows RGB image and ground-truth human mesh. Column third and fourth compare HPS estimation using the BEDLAM-CLIFF \cite{black2023bedlam} and HMR2.0 \cite{goel2023humans} methods. The final column (\name{}-B-CLIFF) presents results obtained by fine-tuning the CLIFF model on the \name{} dataset. These results demonstrate superior performance, particularly in scenarios with heavy occlusion.}
    \vspace{-2ex}
    \label{fig:qualitative}
\end{figure*}

\section{Experiments and Results}
% \vspace{-1ex}
In this section, we highlight the need and significance of our proposed \name{} dataset. We fine-tune the state-of-the-art pose estimation methods using our dataset and report the qualitative and quantitative results. Our results demonstrate improved performance on the HPS estimation task (Section~\ref{subsec: HPS}) using both our proposed synthetic dataset and real-world datasets with occlusions. In Section~\ref{subsec: detection}, we show the enhanced performance of the human object detector and its impact on HPS estimation under occlusion.

\subsection{Human Pose and Shape Estimation}
\label{subsec: HPS}

\noindent \textbf{Dataset and Implementation Setup.} We use approximately 200k and 50k images for training and testing respectively in our proposed \name{} dataset. We fine-tune two versions of state-of-the-art model, CLIFF (trained on Human3.6M \cite{h36m}, MPI-INF-3DHP \cite{MPI-INF-3DHP}, and COCO \cite{mscoco} dataset) and BEDLAM-CLIFF (trained on BEDLAM and AGORA datasets) using our rendered \name{} dataset, resulting in \name{}-CLIFF and \name{}-B-CLIFF, respectively (refer to Figure \ref{fig:yash_train_fig}).
%  name the comparison methods and explain the creation of table1
To benchmark performance, we evaluate on the test split of \name{}, comparing against multiple image and video based baselines, including CLIFF~\cite{li2022cliff}, BEDLAM-HMR~\cite{black2023bedlam}, BEDLAM-CLIFF~\cite{black2023bedlam}, HMR2.0~\cite{goel2023humans}, STRIDE~\cite{stride}, and WHAM~\cite{wham}. 
% We report the mean performance across the cross-validation dataset.
We report the mean performance by evaluating the model using a five-fold cross-validation on our dataset.
%  explain the occlusion levels of our dataset
To quantify occlusion levels, we annotate each 3D joint with a binary occlusion label, marking it as occluded if its corresponding 2D keypoint lies within the ground-truth segmentation mask of the visible human region. Based on the number of visible keypoints (out of 22 total), we categorize images into three occlusion levels: hard occlusion (4-9 visible keypoints), medium occlusion (10-15 visible keypoints), and low occlusion (16-20 visible keypoints).
% explain other datasets

Beyond evaluations on synthetic datasets, we assess the performance of \name{} fine-tuned models on real-world datasets, including 3DPW~\cite{von2018recovering} and OCMotion~\cite{huang2022object}. It is noteworthy that these datasets do not contain significant occlusions. Hence, we introduce two occlusion-augmented variants of 3DPW to evaluate robustness under severe occlusions. In OcclType1-3DPW, we overlay a black patch on a randomly selected keypoint, while in OcclType2-3DPW, we occlude two random keypoints. Further details on occlusion variants of 3DPW and implementation details are provided in the supplementary material.
% We compared our fine-tuned model with other recent image-based HPS estimation methods such as CLIFF, BEDLAM-HMR, BEDLAM-CLIFF, and HMR2.0.        

\noindent \textbf{Results.} We present benchmarking results on the test split of our proposed \name{} dataset in Table~\ref{tab:voccl3d-infer-gt}. Our fine-tuned \name{}-B-CLIFF significantly outperforms existing state-of-the-art image and video-based methods across all occlusion categories. Notably, the performance of all previous methods degrades as occlusion severity increases, highlighting the inherent challenges of HPS estimation under occlusions. We also compare our approach with the plug-and-play method STRIDE \cite{stride}, which performs 3D pose estimation and reports only MPJPE and PA-MPJPE errors. Our results show that STRIDE achieves superior performance when leveraging pseudo-labels from the \name{}-B-CLIFF model compared to those from the original BEDLAM-CLIFF model, further validating the effectiveness of our dataset.
% \calvin{3DPW-AdvOCC also introduces a method that estimates HPS, might be late but can you include those results?}
In Table~\ref{tab:3dpw-infer-gt} and Table~\ref{tab:ocmotion-infer-gt}, we report performance on the 3DPW \cite{3dpw} and OCMotion \cite{ocmotion} datasets, respectively. We observe that both \name{}-B-CLIFF and \name{}-CLIFF outperform all the previous methods, demonstrating their effectiveness in real-world scenarios. However, since these datasets contain minimal occlusions, the improvements over CLIFF remain marginal. To further assess robustness under heavy occlusion, we evaluate on OcclType1-3DPW and OcclType2-3DPW, real-world datasets with significant occlusions. Table~\ref{tab:3dpw-infer-gt} demonstrates that our fine-tuned model achieves substantial improvements, underscoring its robustness in occluded environments. Table~\ref{tab:ocmotion-infer-gt} presents results on the OCMotion dataset, which features lower levels of occlusion. We can observe that \name{}-CLIFF outperforms all comparison methods and demonstrates performance comparable to CLIFF \cite{li2022cliff}. This is expected, as both models are pre-trained on large-scale real-world datasets captured in controlled lab environments similar to OCMotion. Figure~\ref{fig:qualitative} presents the qualitative results on \name{}dataset, showcasing the improvement of our fine-tuned model against other state-of-the-art HPS estimation methods. Additional results are provided in supplementary material. 

% Include the table-1, on the voccl3d test split
% Include the table-2, on the 3DPW and variants

% show the table for OCMotion
\begin{table}[]
\centering
\resizebox{0.8\linewidth}{!}{\begin{tabular}{lccc}
\toprule
\multirow{2}{*}{Method}          & \multicolumn{3}{c}{OCMotion}                     \\
                & MPJPE          & PA-MPJPE       & PVE            \\
\midrule                
CLIFF~\cite{li2022cliff}           & \textbf{64.15} & 40.16          & 79.80          \\
BEDLAM-HMR~\cite{black2023bedlam}      & 73.96          & 41.94          & 92.70          \\
BEDLAM-CLIFF~\cite{black2023bedlam}    & 66.80          & 41.79          & 83.86          \\
HMR2.0~\cite{goel2023humans}          & 69.94          & 43.00          & 87.07          \\
\midrule
VOccl3D-B-CLIFF & 65.96          & 40.59          & 81.96          \\
VOccl3D-CLIFF   & 64.29          & \textbf{39.64} & \textbf{78.56} \\
\bottomrule
\end{tabular}}
\caption{\textbf{Qualitative results of HPS estimation on OCMotion.} Our results show that \name{}-CLIFF outperforms all HPS estimation methods except CLIFF \cite{li2022cliff}. Both CLIFF and \name{}-CLIFF exhibit comparable performance, as both are pre-trained on real-world datasets captured in controlled lab environments, similar to OCMotion \cite{ocmotion}. The best results are in \textbf{bold}.}
\label{tab:ocmotion-infer-gt}
\end{table}

% We reported the numbers of STRIDE with pseudo labels coming from \name{}-B-CLIFF.
% we show the qualitative results and quantitative results across datasets.

\begin{table}[]
% \small
\centering
\resizebox{0.85\linewidth}{!}{
    \begin{tabular}{lcc|cc}
    \toprule
    \multirow{2}{*}{Method} & \multicolumn{2}{c}{3DPW} & \multicolumn{2}{c}{OCMotion} \\
                            & mAP50 & mAP75 & mAP50  & mAP75  \\
    \midrule
    YOLO11                  & 58.99 & 47.14 & 98.84  & 91.80  \\
    VOccl3D-YOLO11          & \textbf{59.89} & \textbf{48.26} & \textbf{99.10}  & \textbf{91.95} \\
    \bottomrule
    \end{tabular}}
\caption{\textbf{Quantitative results of YOLO11 on 3DPW and OCMotion datasets.} The first row presents the human object detection performance of the pre-trained YOLO11 model, while the second row shows the results after fine-tuning with the \name{} dataset. The best results are in \textbf{bold}. We observe an improvement in detection accuracy across both datasets after fine-tuning.}
\vspace{-2ex}
\label{tab:yolo11-detection}
\end{table}

%  HUMAN DETECTION
\subsection{Impact of Human Detector on HPS Estimation}
\label{subsec: detection}

\noindent \textbf{Dataset and Implementation Setup.} Human object detectors play a crucial role in HPS estimation tasks, particularly when inferring from in-the-wild RGB images \cite{li2022cliff, black2023bedlam, stride}. Ideally, HPS estimation performs optimally with ground-truth bounding boxes; however, most human object detectors struggle significantly under occlusion. To address this, we fine-tune the recent YOLO11 detector~\cite{yolo11_ultralytics} on the combined training split of \name{} and MS COCO, referring to the fine-tuned model as \name{}-YOLO11. We show evaluations on 3DPW \cite{3dpw} and OCMotion \cite{ocmotion} datasets.

\noindent \textbf{Results.} Table~\ref{tab:yolo11-detection} compares the detection performance of the pre-trained YOLO11 and \name{}-YOLO11 models on the real-world 3DPW \cite{3dpw} and OCMotion \cite{ocmotion} datasets. Since OCMotion is a single-human dataset with minimal occlusions, the detector achieves high mAP scores. This is likely because most body parts remain visible, making box detection easier. However, due to the presence of multiple individuals in 3DPW, the overall mAP scores remain lower. In short, \name{}-YOLO11 shows significant improvement on datasets with high occlusions. We provide qualitative results in the supplementary material.

% The enhancement is significant on the occluded frames of the dataset depicts the overall marginal improvement of \name{}-YOLO11 in Table~\ref{tab:yolo11-detection}. 

Table~\ref{tab:HPS-yolo11} further evaluates the impact of human detection on HPS estimation. The first, second, and third rows represent the performance of \name{}-CLIFF when human detections are sourced from ground truth, YOLO11, and \name{}-YOLO11, respectively. As expected, the best performance is achieved when using ground-truth detections. However, fine-tuning the detector with \name{} significantly improves HPS estimation compared to detections from the pre-trained YOLO11. These results underscore the critical role of human detection quality in enhancing HPS performance under occlusion.

Our experiments show that fine-tuning an existing HPS estimation model with a large synthetic dataset containing occlusions enhances its performance on both real and synthetic datasets. We observe significant improvements over baseline methods, particularly in scenarios with heavy occlusions. Additionally we demonstrate that the poor results are not solely due to HPS estimations errors but also stem from failures in bounding box detection. The \name{} dataset proves effective in refining pose estimation methods and bounding box predictions under occlusions.

\begin{table}[]
\centering
\resizebox{1.0\linewidth}{!}{
    \begin{tabular}{lcccccc}
    \toprule
    \multirow{2}{*}{Method}                & \multicolumn{3}{c}{3DPW} & \multicolumn{3}{c}{OCMotion} \\
                                           & MPJPE  & PA-MPJPE  & PVE & MPJPE   & PA-MPJPE  & PVE    \\
    \midrule
    \makecell{VOccl3D-CLIFF \\ w/GT}               & 71.10      & 45.98         & 84.25   & 64.29   & 39.64     & 78.56  \\
    \makecell{VOccl3D-CLIFF \\ w/YOLO11}       & 116.52      & 63.35         & 139.74   & 67.16   & 41.30     & 83.00  \\
    \makecell{VOccl3D-CLIFF \\ w/\name{}-YOLO11} & 114.85      & 62.66         & 137.19   & 66.65   & 41.15     & 82.40  \\
    \bottomrule
    \end{tabular}}
\caption{\textbf{3D HPS estimation results using the \name{}-CLIFF model with different bounding box sources.} This table presents HPS estimation performance on the 3DPW and OCMotion datasets. The first row reports results using ground-truth bounding boxes, achieving the highest accuracy. The second and third rows show performance when detections are obtained from the pre-trained YOLO11 and fine-tuned \name{}-YOLO11 models, respectively. Notably, the third row demonstrates a significant improvement over the second, highlighting the impact of detections on HPS estimation.}
\vspace{-2ex}
\label{tab:HPS-yolo11}
\end{table}

%% file: sec/6_conclusion.tex
\vspace{-1ex}
\section{Conclusion}
We introduce \name{}, a novel large-scale video dataset with synthetic humans in real-world scenes under occlusions for 3D human pose and shape estimation. By leveraging a rendering-based approach, \name{} eliminates the need for costly and labor-intensive data collection while providing a diverse and realistic dataset for occlusion-aware research.
Through extensive qualitative and quantitative evaluations, we demonstrate that fine-tuning existing HPS estimation methods with \name{} significantly enhances their performance on real-world datasets with occlusions and on the test split of \name{} dataset. Furthermore, we improve the human object detector, YOLO11 under occluded conditions using \name{}, highlighting the impact of detections in achieving robust HPS estimation in occluded scenarios.
Beyond HPS estimation, \name{} serves as a comprehensive benchmark for evaluating methods across multiple occlusion-aware tasks, including human body-part segmentation, 2D/3D pose estimation, and human bounding box detection. Hence, \name{} sets a new standard for occluded human benchmarks, offering a valuable dataset for advancing occlusion-robust research.

\section*{Acknowledgment} 
\noindent This work was partially supported by USDA NRI grant 2021-67022-33453, NSF grants CMMI-2326309 and CNS-2312395.  This research was also partially supported by the Office of the Director of National Intelligence (ODNI), specifically through the Intelligence Advanced Research Projects Activity (IARPA), under contract number [2022- 21102100007]. The views and conclusions in this research reflect those of the authors and should not be construed as officially representing the policies, whether explicitly or implicitly, of ODNI, IARPA, or the U.S. Government. Nevertheless, the U.S. Government retains the authorization to reproduce and distribute reprints for official government purposes, regardless of any copyright notices included.

%% file: suppl_under_main.tex
% \begin{document}
\renewcommand{\thesection}{\Alph{section}}
\setcounter{section}{0}

\begin{center}
    \Large \textbf{Supplementary Material}
\end{center}
\section{Implementation details.}
\subsection{Human Pose and Shape estimation.}
We fine-tune CLIFF~\cite{li2022cliff} and BEDLAM-CLIFF~\cite{black2023bedlam} for HPS estimation using approximately 200k images from our \name{} dataset. CLIFF is trained on real 2D datasets such as COCO~\cite{mscoco} and MPII~\cite{andriluka20142d}, as well as 3D datasets like Human3.6M~\cite{ionescu2013human3} and 3DHP~\cite{MPI-INF-3DHP}, while BEDLAM-CLIFF is originally trained on synthetic datasets such as BEDLAM~\cite{black2023bedlam} and AGORA~\cite{patel2021agora}.
We fine-tune these models on a single NVIDIA GeForce RTX 3090 Ti GPU. We adopt hyperparameters and loss functions from~\cite{black2023bedlam} for fine-tuning. We optimize the models using the Adam optimizer with a learning rate of 0.00005 and zero weight decay. To prevent overfitting, we employ early stopping. We use a batch size of 64 and resize input images to $224 \times 224$ dimension.

We report errors after converting SMPL-X bodies to SMPL using a pre-trained joint regressor mapping and aligning the pelvis of these bodies. We evaluate CLIFF, BEDLAM-CLIFF, BEDLAM-HMR, HMR2.0, WHAM, and STRIDE by re-running their evaluations using the official code repositories.

\begin{figure*}[!ht]
    \centering
    \includegraphics[width=0.9\linewidth]{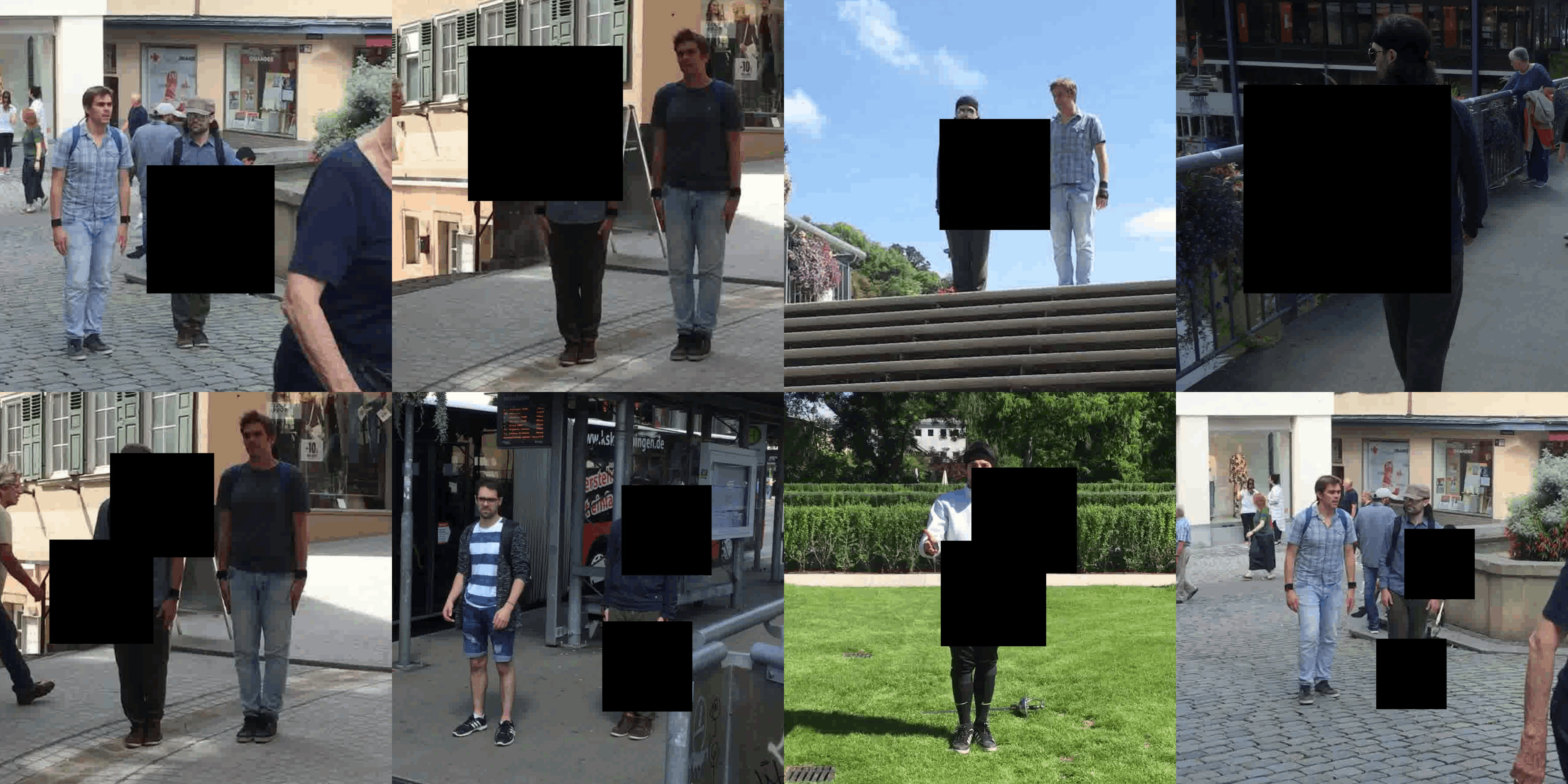}
    \caption{Samples of OcclType1-3DPW (top row) and OcclType2-3DPW (bottom row) dataset.}
    \label{fig:3dpw_variants}
\end{figure*}

We create two variants of the 3DPW dataset, OcclType1-3DPW and OcclType2-3DPW, by overlaying black patches to evaluate performance on highly occluded real-world datasets. OcclType1-3DPW is generated by randomly adding a black patch over a single 2D keypoint from the 22 openpose joints, while OcclType2-3DPW contains images with two black patches placed on random 2D keypoints. The added patches are square-shaped, with dimensions covering 60\% of the human height in OcclType1-3DPW and 40\% of the human height in OcclType2-3DPW.
Figure~\ref{fig:3dpw_variants} illustrates sample images from OcclType1-3DPW and OcclType2-3DPW. We follow the same evaluation procedure for real-world datasets, including 3DPW, OcclType1-3DPW, OcclType2-3DPW, and OCMotion, as we do for the \name{} dataset.

\noindent \textbf{Evaluation metrics.} Following prior works, we use standard metrics to report the performance of human pose and shape estimation. MPJPE and PVE represent the average error in joints and vertices respectively after aligning the pelvis. PA-MPJPE reports the average error after aligning the rotation and scale. All errors are in mm.

\begin{figure}[!ht]
    \centering
    \includegraphics[width=0.9\linewidth]{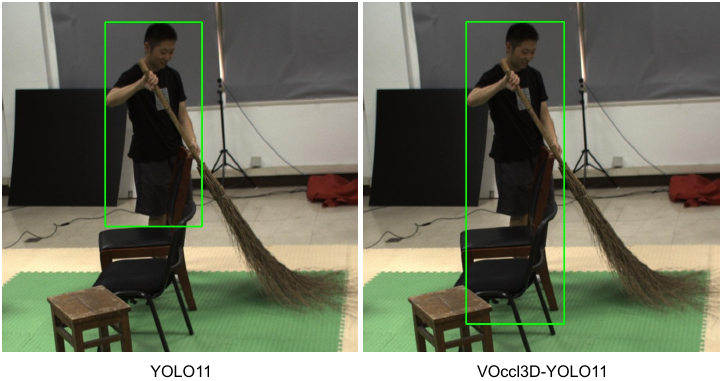}
    \caption{\textbf{Human detection under occlusion on OCMotion using YOLO11.} The left image illustrates detection performance with the pre-trained YOLO11, while the right image shows improved detection after fine-tuning YOLO11 with the \name{} dataset, resulting in \name{}-YOLO11.}
    \label{fig:detection}
\end{figure}

\subsection{Human detector.}
We conduct our experiments on the YOLO11 detector using the official Ultralytics codebase~\cite{yolo11_ultralytics}. The original YOLO11 model is pre-trained on the MS COCO dataset~\cite{mscoco}. To enhance its performance under occlusions, we fine-tune YOLO11 on the combined train split of \name{} and MS COCO, resulting in \name{}-YOLO11. We fine-tune the model for 50 epochs with a batch size of 32 on a single NVIDIA GeForce RTX 3090 Ti GPU. Following~\cite{yolo11_ultralytics}, we resize input images to $640\times640$ and train using a learning rate of 0.01 with a weight decay of 0.0005. Additionally, we set the loss function weights to 7.5 for the bounding box component and 0.5 for the classification component to optimize detection performance.

Figure~\ref{fig:detection} shows the qualitative performance of YOLO11 and \name{}-YOLO11, where we show an improved performance of \name{}-YOLO11 under high occlusions. 

\noindent \textbf{Evaluation metrics.}We evaluate detector performance using mean Average Precision (mAP) at Intersection over Union (IoU) thresholds of 0.50 and 0.75, referred to as mAP50 and mAP75, respectively. Unlike standard bounding box labels that include only visible human regions, we provide bounding box annotations that cover the entire human body, including both visible and occluded parts.

\section{Additional related works.}
\noindent\textbf{Datasets for Pose Estimation}
Previous works have proposed several datasets for HPSE, which are either video-based or image-based. One of the pioneers in this field is the CMU Motion Capture dataset which primarily contained 3D skeletal data without RGB images. This dataset included a wide range of activities like dancing, walking, and sports and served as a cornerstone for tasks like animation, pose estimation, and gaming. Further, in 2016, the MSCOCO dataset \cite{mscoco} was released which initially contained over 200,000 labeled images covering 80 object categories, including humans. The scale of this dataset provided a wealth of data that was unprecedented for pose estimation tasks at the time. Additionally, MSCOCO introduced keypoint annotations for human pose estimation, providing 17 key points per person. The Archive of Motion Capture As Surface Shapes (AMASS) dataset \cite{AMASS}, introduced in \cite{AMASS}, is a large human motion database that unifies various optical marker-based motion capture datasets under a common framework and parameterization. This dataset contains 40 hours of human motion data, spanning over 300 subjects, and motivated large-scale pre-training in a variety of follow-up HPS works \cite{MPT, multi-layer, kocabas2019vibe, motionbert2022}. The recent 3D Poses in the Wild (3DPW) dataset \cite{3dpw} is a widely-used benchmark for evaluating 3D human pose estimation methods in natural, unstructured environments, providing accurate 3D pose annotations derived from synchronized video and inertial measurement unit (IMU) data. This dataset comprises over 51,000 frames and across 60 video sequences. Although these datasets fueled the state-of-the-art methods but contain limited occlusions in their samples. This makes methods trained on these datasets vulnerable to occlusions, limiting their ability to generalize to unseen scenarios with significant occlusions.

\section{Qualitative examples}
\begin{figure*}[!ht]
    \centering
    \includegraphics[width=\textwidth]{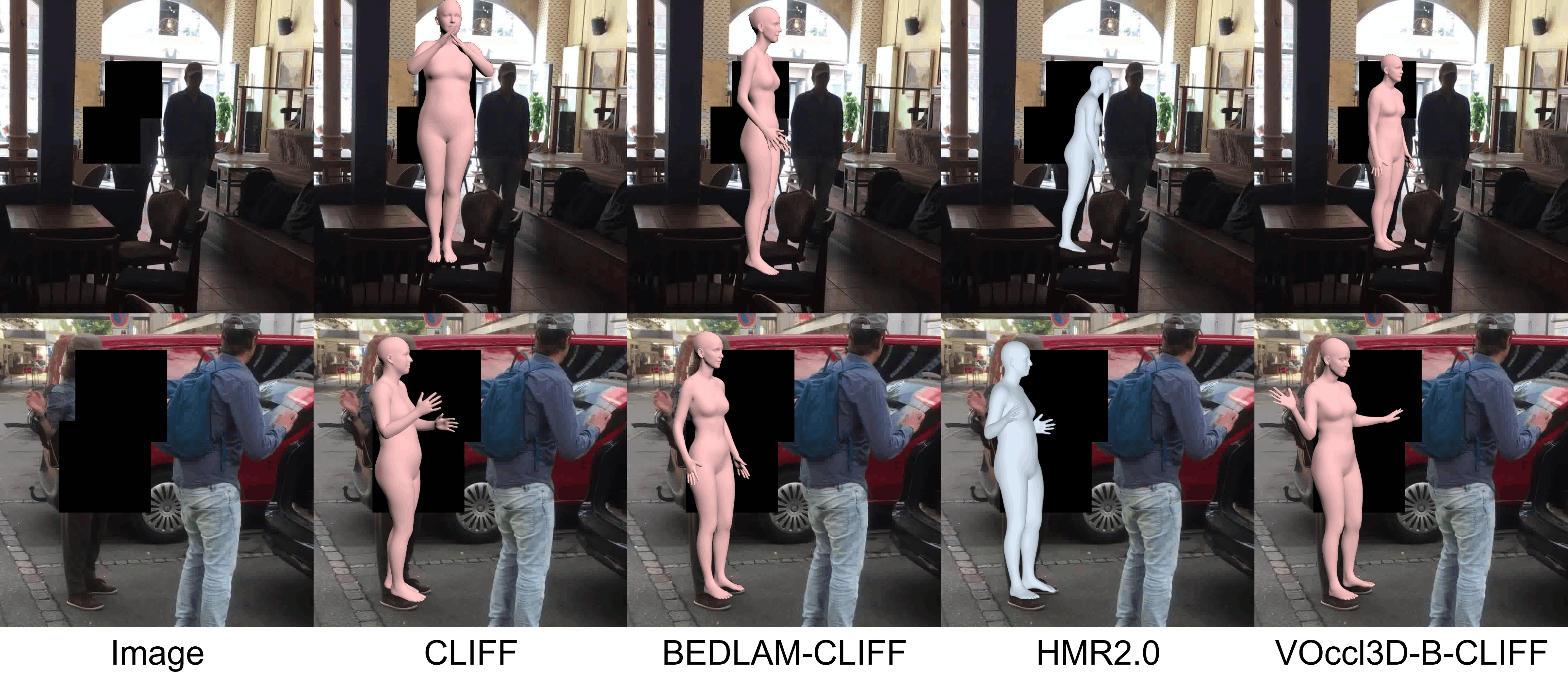}
    \caption{\textbf{Qualitative comparison of HPS estimation methods on OcclType2-3DPW dataset.} Column 1 represents input RGB image. Columns 2–4 compare HPS estimation using the CLIFF \cite{li2022cliff}, BEDLAM-CLIFF \cite{black2023bedlam}, and HMR2.0 \cite{goel2023humans} methods. The final column (\name{}-B-CLIFF) presents results obtained by fine-tuning the CLIFF model on the \name{} dataset.}
    \label{fig:qualitative_3dpw}
\end{figure*}

\begin{figure*}[!ht]
    \centering
    \includegraphics[width=\textwidth]{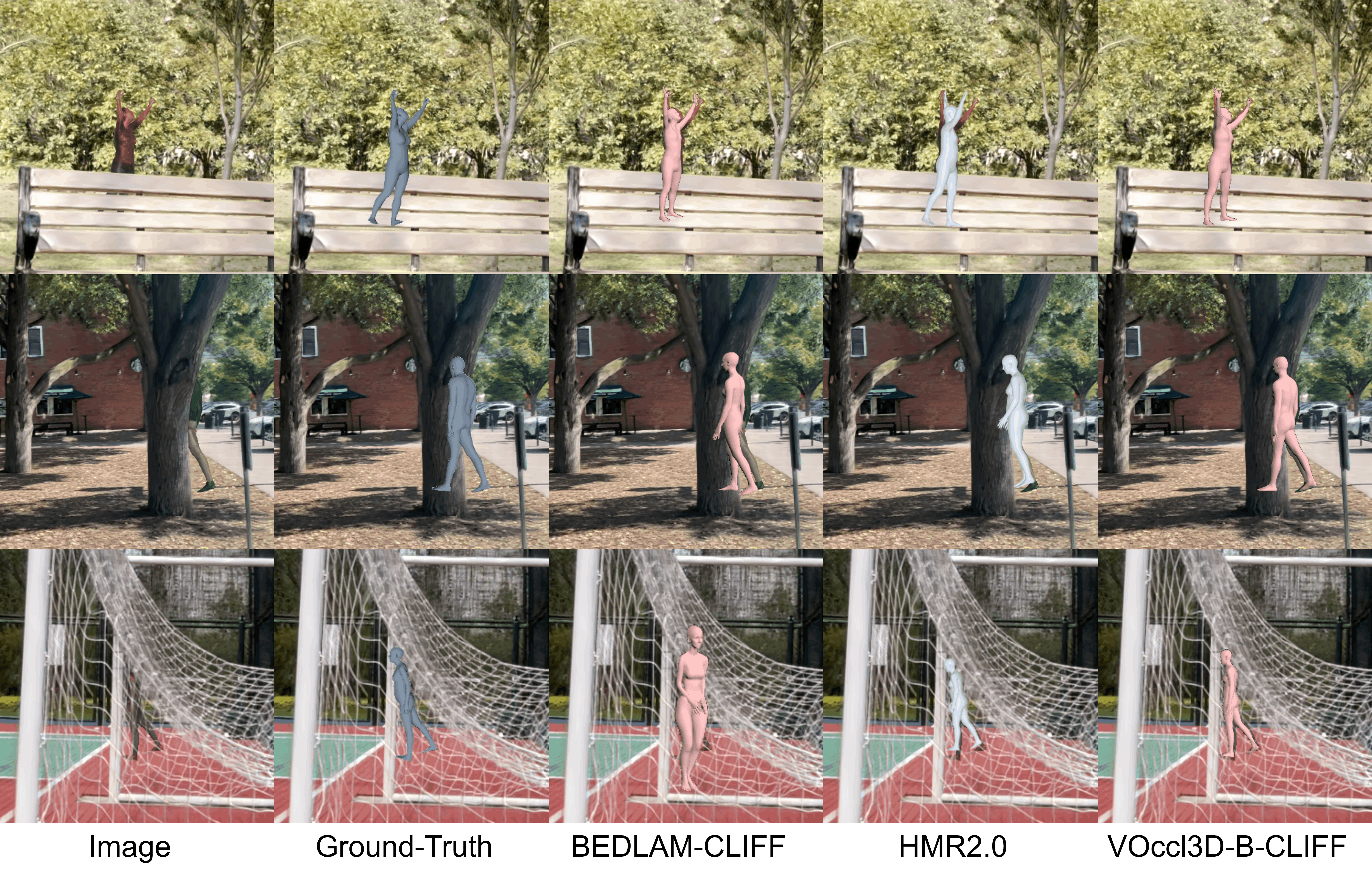}
    \caption{\textbf{Qualitative comparison of HPS estimation methods on \name{} dataset.} Column 1 and 2 represents input RGB image and ground truth pose. Columns 3 and 4 compare HPS estimation using the BEDLAM-CLIFF \cite{black2023bedlam}, and HMR2.0 \cite{goel2023humans} methods. The final column (\name{}-B-CLIFF) presents results obtained by fine-tuning the CLIFF model on the \name{} dataset.}
    \label{fig:qualitative_supp}
\end{figure*}

\begin{figure*}[!ht]
    \centering
    \includegraphics[width=0.9\textwidth]{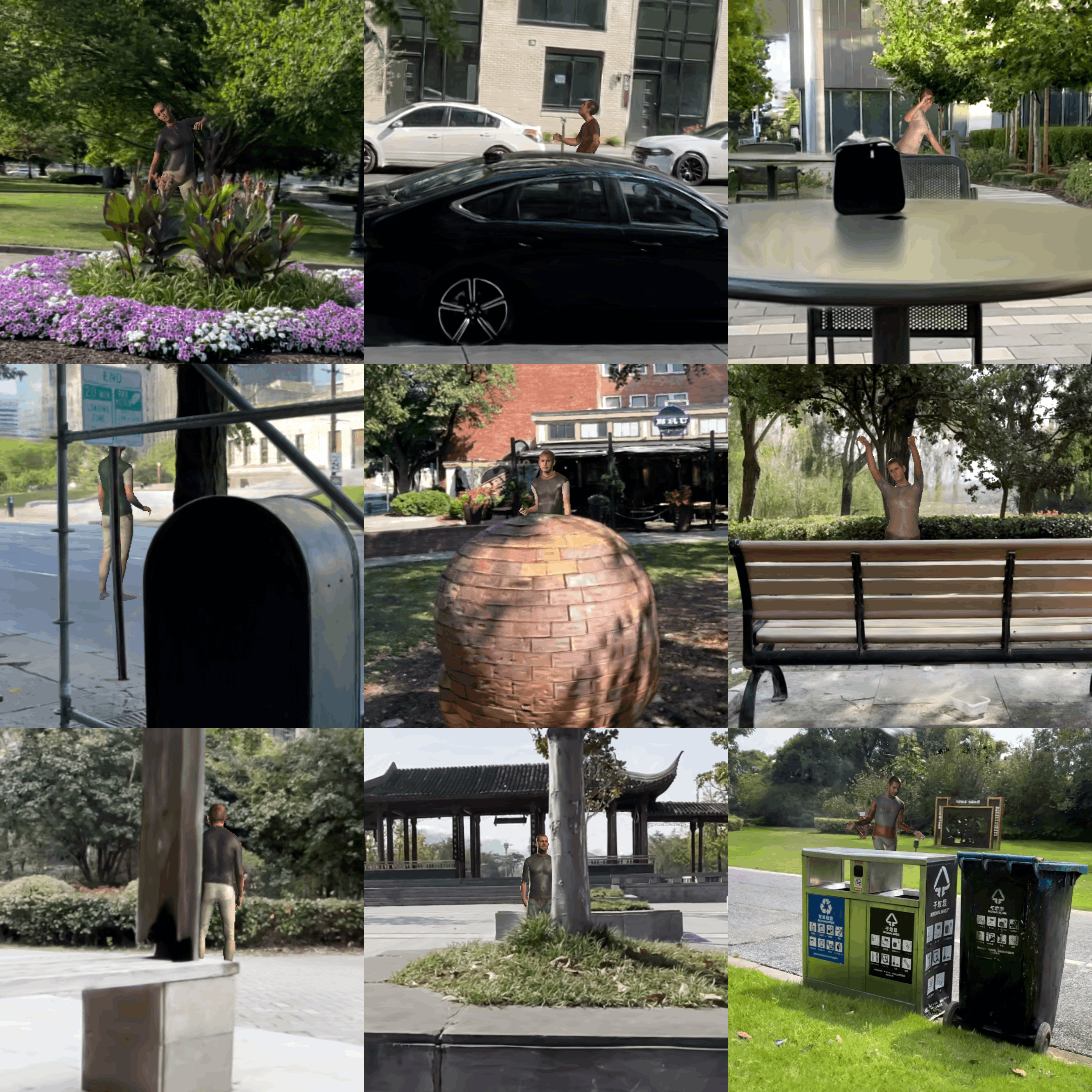}
    \caption{\textbf{Samples of \name{} dataset.} The samples from \name{} dataset illustrates various diversity in real occlusions, human motions, and clothing textures.}
    \label{fig:voccl3d_sample}
\end{figure*}

In this section, we present the qualitative results of our fine-tuned model, \name{}-B-CLIFF, in comparison with other HPS estimation methods. Figure~\ref{fig:qualitative_3dpw} illustrates qualitative results on the OcclType2-3DPW dataset, while Figure~\ref{fig:qualitative_supp} provides additional qualitative comparisons on the test split of \name{}. We observe the superior performance of \name{}-B-CLIFF across multiple datasets. Additionally, Figure~\ref{fig:voccl3d_sample} showcases further sample images from the \name{} dataset.

\begin{table*}[t]
% \begin{wraptable}[l]{1\columnwidth}
    % \vspace{-2em}
    \centering
    \resizebox{1\linewidth}{!}{
    \begin{tabular}{llllllllll}
    \toprule
    Datasets           & \#Sub             & \#Frames           & Image           & Subj/image & Motion           & Ground-Truth & Occlusion & Multi-level Occlusion & Video data \\
    \midrule
    SURREAL            & 145               & $\sim$6.5M         & composite       & 1          & \textgreater{}2k & SMPL         & No        & No                    & No         \\
    MPI-INF-3DHP-Train & 8                 & \textgreater{}1.3M & mixed/composite & 1          & 8+               & 3D joints    & No        & No                    & Yes          \\
    AGORA              & \textgreater{}350 & $\sim$18k          & rendered        & 5-15       & n/a              & SMPL-X       & Yes       & No                    & No         \\
    BEDLAM             & 217               & 380k               & rendered        & 1-10       & 2311             & SMPL-X       & No        & No                    & Yes        \\
    SynthMoCap         & $\sim$200         & $\sim$100k         & rendered        & 1-4        & n/a              & SMPL-X       & No        & No                    & No         \\
    OCMotion           & 8                 & 300k               & captured        & 1          & 43               & SMPL         & Yes       & No                    & Yes        \\
    \midrule
    \textbf{\name}     & $\sim$200         & $\sim$250k         & rendered        & 1          & 400              & SMPL-X       & Yes       & Yes                   & Yes       \\
    \bottomrule \\
    \end{tabular}
    }
    % \vspace{-1em}
    \caption{Comparison of synthetic datasets and real dataset with occlusion for 3D human pose estimation.}
    % \vspace{-2em}
    \label{tab:dataset_summary}
\end{table*}

% adding the limitation section
\input{sec/5_limitation}

% \newpage
% \vspace*{40cm}
% \setlength\bibitemsep{0pt}
% \clearpage
% {
%     \small
%     \bibliographystyle{ieeenat_fullname}
%     \bibliography{main}
% }

% \end{document}

%% file: sec/5_limitation.tex
\section{Limitations and Future Work}
Our work highlights the need and importance of a large-scale, realistic occluded human dataset for performing the task of human pose and shape estimation. By releasing this dataset and the associated tools for repopulation, we aim to enable the research community to systematically evaluate their algorithms under challenging occlusion scenarios.

% our dataset is limited by a set of diverse background scenes, human motions, and clothing texture, which can be extended to create a vast diverse dataset. 
Currently, the visual quality of our synthetic humans is limited by the lack of open-source high-fidelity assets, such as garments, hairstyles, footwear, and diverse human motions, which are constrained by the AMASS dataset. Moreover, our rendering pipeline relies on predefined camera poses to generate images with substantial occlusions. A promising direction for future work would be to develop an end-to-end framework that can automatically generate occlusion-rich sequences without requiring externally provided camera parameters.

% We have shown the evaluation on existing HPS estimation methods to claim that they fail to perform well under occlusion. However, there can be multiple plausible poses for the occluded human body, which makes it difficult for any pose estimation algorithm to generalize under the occlusion. For future work an interesting problem would be to predict multiple plausible poses for the occluded human body.
Although the VOccl3D dataset offers realistic occlusion scenarios, a noticeable gap remains between synthetic and real-world data. Bridging this sim-to-real gap represents an important avenue for future research in realistic human pose estimation. Additionally, our dataset holds potential utility for broader research efforts focused on occlusion-aware learning across various modalities, including human silhouette extraction, body-part segmentation, 2D keypoint estimation, and bounding box detection

%% file: main.bbl
\begin{thebibliography}{62}
\providecommand{\natexlab}[1]{#1}
\providecommand{\url}[1]{\texttt{#1}}
\expandafter\ifx\csname urlstyle\endcsname\relax
  \providecommand{\doi}[1]{doi: #1}\else
  \providecommand{\doi}{doi: \begingroup \urlstyle{rm}\Url}\fi

\bibitem[Andriluka et~al.(2014)Andriluka, Pishchulin, Gehler, and Schiele]{andriluka20142d}
Mykhaylo Andriluka, Leonid Pishchulin, Peter Gehler, and Bernt Schiele.
\newblock 2d human pose estimation: New benchmark and state of the art analysis.
\newblock In \emph{Proceedings of the IEEE Conference on computer Vision and Pattern Recognition}, pages 3686--3693, 2014.

\bibitem[Arnab et~al.(2019)Arnab, Doersch, and Zisserman]{Arnab_CVPR_2019}
Anurag Arnab, Carl Doersch, and Andrew Zisserman.
\newblock Exploiting temporal context for 3d human pose estimation in the wild.
\newblock In \emph{Computer Vision and Pattern Recognition (CVPR)}, 2019.

\bibitem[Bauer et~al.(2023)Bauer, Bouazizi, Kressel, and Flohr]{pose_aut2}
Peter Bauer, Arij Bouazizi, Ulrich Kressel, and Fabian~B. Flohr.
\newblock Weakly supervised multi-modal 3d human body pose estimation for autonomous driving.
\newblock In \emph{2023 IEEE Intelligent Vehicles Symposium (IV)}, pages 1--7, 2023.

\bibitem[Black et~al.(2023)Black, Patel, Tesch, and Yang]{black2023bedlam}
Michael~J. Black, Priyanka Patel, Joachim Tesch, and Jinlong Yang.
\newblock Bedlam: A synthetic dataset of bodies exhibiting detailed lifelike animated motion, 2023.

\bibitem[Casas and Comino-Trinidad(2023)]{casas2023smplitex}
Dan Casas and Marc Comino-Trinidad.
\newblock {SMPLitex: A Generative Model and Dataset for 3D Human Texture Estimation from Single Image}.
\newblock In \emph{British Machine Vision Conference (BMVC)}, 2023.

\bibitem[Cheng et~al.(2019)Cheng, Yang, Wang, Wending, and Tan]{occ-aware}
Yu Cheng, Bo Yang, Bo Wang, Yan Wending, and Robby Tan.
\newblock Occlusion-aware networks for 3d human pose estimation in video.
\newblock In \emph{2019 IEEE/CVF International Conference on Computer Vision (ICCV)}, pages 723--732, 2019.

\bibitem[Clever et~al.(2020)Clever, Erickson, Kapusta, Turk, Liu, and Kemp]{clever2020bodies}
Henry~M Clever, Zackory Erickson, Ariel Kapusta, Greg Turk, Karen Liu, and Charles~C Kemp.
\newblock Bodies at rest: 3d human pose and shape estimation from a pressure image using synthetic data.
\newblock In \emph{Proceedings of the IEEE/CVF conference on computer vision and pattern recognition}, pages 6215--6224, 2020.

\bibitem[Cormier et~al.(2022)Cormier, Clepe, Specker, and Beyerer]{surveillance1}
Mickael Cormier, Aris Clepe, Andreas Specker, and Jürgen Beyerer.
\newblock Where are we with human pose estimation in real-world surveillance?
\newblock In \emph{2022 IEEE/CVF Winter Conference on Applications of Computer Vision Workshops (WACVW)}, pages 591--601, 2022.

\bibitem[Cotton(2022)]{clinical_2}
R~James Cotton.
\newblock Posepipe: Open-source human pose estimation pipeline for rehabilitation research.
\newblock \emph{Archives of Physical Medicine and Rehabilitation}, 103\penalty0 (12):\penalty0 e161--e162, 2022.

\bibitem[Cui and Kang(2023)]{gait_1}
Yufeng Cui and Yimei Kang.
\newblock Multi-modal gait recognition via effective spatial-temporal feature fusion.
\newblock In \emph{2023 IEEE/CVF Conference on Computer Vision and Pattern Recognition (CVPR)}, pages 17949--17957, 2023.

\bibitem[Fu et~al.(2023)Fu, Meng, Hou, Hu, and Huang]{gait_2}
Yang Fu, Shibei Meng, Saihui Hou, Xuecai Hu, and Yongzhen Huang.
\newblock Gpgait: Generalized pose-based gait recognition.
\newblock In \emph{2023 IEEE/CVF International Conference on Computer Vision (ICCV)}, pages 19538--19547, 2023.

\bibitem[Geng et~al.(2023)Geng, Yang, Jiao, Zeng, and Chen]{multi-layer}
Lei Geng, Wenzhu Yang, Yanyan Jiao, Shuang Zeng, and Xinting Chen.
\newblock A multilayer human motion prediction perceptron by aggregating repetitive motion.
\newblock \emph{Mach. Vision Appl.}, 34\penalty0 (6), 2023.

\bibitem[Goel et~al.(2023)Goel, Pavlakos, Rajasegaran, Kanazawa, and Malik]{goel2023humans}
Shubham Goel, Georgios Pavlakos, Jathushan Rajasegaran, Angjoo Kanazawa, and Jitendra Malik.
\newblock Humans in 4d: Reconstructing and tracking humans with transformers.
\newblock In \emph{Proceedings of the IEEE/CVF International Conference on Computer Vision}, pages 14783--14794, 2023.

\bibitem[Hewitt et~al.(2024)Hewitt, Saleh, Aliakbarian, Petikam, Rezaeifar, Florentin, Hosenie, Cashman, Valentin, Cosker, et~al.]{hewitt2024look}
Charlie Hewitt, Fatemeh Saleh, Sadegh Aliakbarian, Lohit Petikam, Shideh Rezaeifar, Louis Florentin, Zafiirah Hosenie, Thomas~J Cashman, Julien Valentin, Darren Cosker, et~al.
\newblock Look ma, no markers: Holistic performance capture without the hassle.
\newblock \emph{ACM Transactions on Graphics}, 43\penalty0 (6):\penalty0 1--12, 2024.

\bibitem[Huang et~al.(2022)Huang, Zhang, and Wang]{huang2022object}
Buzhen Huang, Tianshu Zhang, and Yangang Wang.
\newblock Object-occluded human shape and pose estimation with probabilistic latent consistency.
\newblock \emph{IEEE Transactions on Pattern Analysis and Machine Intelligence}, 45\penalty0 (4):\penalty0 5010--5026, 2022.

\bibitem[Huang et~al.(2023)Huang, Zhang, and Wang]{ocmotion}
Buzhen Huang, Tianshu Zhang, and Yangang Wang.
\newblock Object-occluded human shape and pose estimation with probabilistic latent consistency.
\newblock \emph{IEEE Transactions on Pattern Analysis and Machine Intelligence}, 45\penalty0 (4):\penalty0 5010--5026, 2023.

\bibitem[Ionescu et~al.(2013)Ionescu, Papava, Olaru, and Sminchisescu]{ionescu2013human3}
Catalin Ionescu, Dragos Papava, Vlad Olaru, and Cristian Sminchisescu.
\newblock Human3. 6m: Large scale datasets and predictive methods for 3d human sensing in natural environments.
\newblock \emph{IEEE transactions on pattern analysis and machine intelligence}, 36\penalty0 (7):\penalty0 1325--1339, 2013.

\bibitem[Ionescu et~al.(2014)Ionescu, Papava, Olaru, and Sminchisescu]{h36m}
Catalin Ionescu, Dragos Papava, Vlad Olaru, and Cristian Sminchisescu.
\newblock Human3.6m: Large scale datasets and predictive methods for 3d human sensing in natural environments.
\newblock \emph{IEEE Transactions on Pattern Analysis and Machine Intelligence}, 36\penalty0 (7):\penalty0 1325--1339, 2014.

\bibitem[Jocher and Qiu(2024)]{yolo11_ultralytics}
Glenn Jocher and Jing Qiu.
\newblock Ultralytics yolo11, 2024.

\bibitem[Juraev et~al.(2022)Juraev, Ghimire, Alikhanov, Kakani, and Kim]{surveillance2}
Sardor Juraev, Akash Ghimire, Jumabek Alikhanov, Vijay Kakani, and Hakil Kim.
\newblock Exploring human pose estimation and the usage of synthetic data for elderly fall detection in real-world surveillance.
\newblock \emph{IEEE Access}, 10:\penalty0 94249--94261, 2022.

\bibitem[Kanazawa et~al.(2017)Kanazawa, Black, Jacobs, and Malik]{Kanazawa2017EndtoEndRO}
Angjoo Kanazawa, Michael~J. Black, David~W. Jacobs, and Jitendra Malik.
\newblock End-to-end recovery of human shape and pose.
\newblock \emph{2018 IEEE/CVF Conference on Computer Vision and Pattern Recognition}, pages 7122--7131, 2017.

\bibitem[Kerbl et~al.(2023)Kerbl, Kopanas, Leimk{\"u}hler, and Drettakis]{3dgs}
Bernhard Kerbl, Georgios Kopanas, Thomas Leimk{\"u}hler, and George Drettakis.
\newblock 3d gaussian splatting for real-time radiance field rendering.
\newblock \emph{ACM Transactions on Graphics}, 42\penalty0 (4), 2023.

\bibitem[Kim and Chang(2021)]{attn-seq}
Do-Yeop Kim and Ju-Yong Chang.
\newblock Attention-based 3d human pose sequence refinement network.
\newblock \emph{Sensors}, 21\penalty0 (13), 2021.

\bibitem[Kocabas et~al.(2020)Kocabas, Athanasiou, and Black]{kocabas2019vibe}
Muhammed Kocabas, Nikos Athanasiou, and Michael~J. Black.
\newblock Vibe: Video inference for human body pose and shape estimation.
\newblock In \emph{The IEEE Conference on Computer Vision and Pattern Recognition (CVPR)}, 2020.

\bibitem[Kocabas et~al.(2021)Kocabas, Huang, Hilliges, and Black]{Kocabas_PARE_2021}
Muhammed Kocabas, Chun-Hao~P. Huang, Otmar Hilliges, and Michael~J. Black.
\newblock {PARE}: Part attention regressor for {3D} human body estimation.
\newblock In \emph{Proc. International Conference on Computer Vision (ICCV)}, pages 11127--11137, 2021.

\bibitem[Koleini et~al.(2025)Koleini, Saleem, Wang, Xue, Helmy, and Fenwick]{koleini2025biopose}
Farnoosh Koleini, Muhammad~Usama Saleem, Pu Wang, Hongfei Xue, Ahmed Helmy, and Abbey Fenwick.
\newblock Biopose: Biomechanically-accurate 3d pose estimation from monocular videos.
\newblock In \emph{2025 IEEE/CVF Winter Conference on Applications of Computer Vision (WACV)}, pages 6330--6339. IEEE, 2025.

\bibitem[Lal et~al.(2024)Lal, Bachu, Garg, Dutta, Ta, Raychaudhuri, Cruz, Asif, and Roy-Chowdhury]{stride}
Rohit Lal, Saketh Bachu, Yash Garg, Arindam Dutta, Calvin-Khang Ta, Dripta~S. Raychaudhuri, Hannah~Dela Cruz, M.~Salman Asif, and Amit~K. Roy-Chowdhury.
\newblock Stride: Single-video based temporally continuous occlusion robust 3d pose estimation, 2024.

\bibitem[Li et~al.(2022)Li, Liu, Zhang, Xu, and Yan]{li2022cliff}
Zhihao Li, Jianzhuang Liu, Zhensong Zhang, Songcen Xu, and Youliang Yan.
\newblock Cliff: Carrying location information in full frames into human pose and shape estimation, 2022.

\bibitem[Lin et~al.(2022)Lin, Lin, Liang, Liu, and Wang]{MPT}
Kevin Lin, Chung-Ching Lin, Lin Liang, Zicheng Liu, and Lijuan Wang.
\newblock Mpt: Mesh pre-training with transformers for human pose and mesh reconstruction.
\newblock \emph{2024 IEEE/CVF Winter Conference on Applications of Computer Vision (WACV)}, pages 3403--3413, 2022.

\bibitem[Lin et~al.(2014)Lin, Maire, Belongie, Hays, Perona, Ramanan, Doll{\'a}r, and Zitnick]{mscoco}
Tsung-Yi Lin, Michael Maire, Serge Belongie, James Hays, Pietro Perona, Deva Ramanan, Piotr Doll{\'a}r, and C.~Lawrence Zitnick.
\newblock Microsoft coco: Common objects in context.
\newblock In \emph{Computer Vision -- ECCV 2014}, pages 740--755, Cham, 2014. Springer International Publishing.

\bibitem[Ling et~al.(2024)Ling, Sheng, Tu, Zhao, Xin, Wan, Yu, Guo, Yu, Lu, et~al.]{dl3dv}
Lu Ling, Yichen Sheng, Zhi Tu, Wentian Zhao, Cheng Xin, Kun Wan, Lantao Yu, Qianyu Guo, Zixun Yu, Yawen Lu, et~al.
\newblock Dl3dv-10k: A large-scale scene dataset for deep learning-based 3d vision.
\newblock In \emph{Proceedings of the IEEE/CVF Conference on Computer Vision and Pattern Recognition}, pages 22160--22169, 2024.

\bibitem[Loper et~al.(2015)Loper, Mahmood, Romero, Pons-Moll, and Black]{SMPL}
Matthew Loper, Naureen Mahmood, Javier Romero, Gerard Pons-Moll, and Michael~J. Black.
\newblock {SMPL}: A skinned multi-person linear model.
\newblock \emph{ACM Trans. Graphics (Proc. SIGGRAPH Asia)}, 34\penalty0 (6):\penalty0 248:1--248:16, 2015.

\bibitem[Luo et~al.(2020)Luo, Golestaneh, and Kitani]{meva}
Zhengyi Luo, S.~Alireza Golestaneh, and Kris~M. Kitani.
\newblock 3d human motion estimation via motion compression and refinement.
\newblock In \emph{Proceedings of the Asian Conference on Computer Vision (ACCV)}, 2020.

\bibitem[Mahmood et~al.(2019)Mahmood, Ghorbani, F.~Troje, Pons-Moll, and Black]{AMASS}
Naureen Mahmood, Nima Ghorbani, Nikolaus F.~Troje, Gerard Pons-Moll, and Michael~J. Black.
\newblock Amass: Archive of motion capture as surface shapes.
\newblock In \emph{The IEEE International Conference on Computer Vision (ICCV)}, 2019.

\bibitem[Mehta et~al.(2017)Mehta, Rhodin, Casas, Fua, Sotnychenko, Xu, and Theobalt]{MPI-INF-3DHP}
Dushyant Mehta, Helge Rhodin, Dan Casas, Pascal Fua, Oleksandr Sotnychenko, Weipeng Xu, and Christian Theobalt.
\newblock Monocular 3d human pose estimation in the wild using improved cnn supervision.
\newblock In \emph{2017 International Conference on 3D Vision (3DV)}, pages 506--516, 2017.

\bibitem[Mildenhall et~al.(2021)Mildenhall, Srinivasan, Tancik, Barron, Ramamoorthi, and Ng]{mildenhall2021nerf}
Ben Mildenhall, Pratul~P Srinivasan, Matthew Tancik, Jonathan~T Barron, Ravi Ramamoorthi, and Ren Ng.
\newblock Nerf: Representing scenes as neural radiance fields for view synthesis.
\newblock \emph{Communications of the ACM}, 65\penalty0 (1):\penalty0 99--106, 2021.

\bibitem[Moccia et~al.(2019)Moccia, Migliorelli, Carnielli, and Frontoni]{clinical_1}
Sara Moccia, Lucia Migliorelli, Virgilio Carnielli, and Emanuele Frontoni.
\newblock Preterm infants’ pose estimation with spatio-temporal features.
\newblock \emph{IEEE Transactions on Biomedical Engineering}, 67\penalty0 (8):\penalty0 2370--2380, 2019.

\bibitem[Monszpart et~al.(2019)Monszpart, Guerrero, Ceylan, Yumer, and Mitra]{i3db}
Aron Monszpart, Paul Guerrero, Duygu Ceylan, Ersin Yumer, and Niloy~J. Mitra.
\newblock imapper: interaction-guided scene mapping from monocular videos.
\newblock \emph{ACM Trans. Graph.}, 38\penalty0 (4), 2019.

\bibitem[Patel et~al.(2021)Patel, Huang, Tesch, Hoffmann, Tripathi, and Black]{patel2021agora}
Priyanka Patel, Chun-Hao~P Huang, Joachim Tesch, David~T Hoffmann, Shashank Tripathi, and Michael~J Black.
\newblock Agora: Avatars in geography optimized for regression analysis.
\newblock In \emph{Proceedings of the IEEE/CVF Conference on Computer Vision and Pattern Recognition}, pages 13468--13478, 2021.

\bibitem[Pavlakos et~al.(2019)Pavlakos, Choutas, Ghorbani, Bolkart, Osman, Tzionas, and Black]{pavlakos2019expressive}
Georgios Pavlakos, Vasileios Choutas, Nima Ghorbani, Timo Bolkart, Ahmed~AA Osman, Dimitrios Tzionas, and Michael~J Black.
\newblock Expressive body capture: 3d hands, face, and body from a single image.
\newblock In \emph{Proceedings of the IEEE/CVF conference on computer vision and pattern recognition}, pages 10975--10985, 2019.

\bibitem[Pavllo et~al.(2018)Pavllo, Feichtenhofer, Grangier, and Auli]{Pavllo20183DHP}
Dario Pavllo, Christoph Feichtenhofer, David Grangier, and Michael Auli.
\newblock 3d human pose estimation in video with temporal convolutions and semi-supervised training.
\newblock \emph{2019 IEEE/CVF Conference on Computer Vision and Pattern Recognition (CVPR)}, pages 7745--7754, 2018.

\bibitem[Purkr{\'a}bek and Matas(2023)]{purkrabek2023improving}
Miroslav Purkr{\'a}bek and Ji{\v{r}}{\'\i} Matas.
\newblock Improving 2d human pose estimation across unseen camera views with synthetic data.
\newblock \emph{arXiv preprint arXiv:2307.06737}, 2023.

\bibitem[Ramamoorthi and Hanrahan(2001)]{ramamoorthi2001efficient}
Ravi Ramamoorthi and Pat Hanrahan.
\newblock An efficient representation for irradiance environment maps.
\newblock In \emph{Proceedings of the 28th annual conference on Computer graphics and interactive techniques}, pages 497--500, 2001.

\bibitem[Rempe et~al.(2021)Rempe, Birdal, Hertzmann, Yang, Sridhar, and Guibas]{rempe2021humor}
Davis Rempe, Tolga Birdal, Aaron Hertzmann, Jimei Yang, Srinath Sridhar, and Leonidas~J. Guibas.
\newblock Humor: 3d human motion model for robust pose estimation.
\newblock In \emph{International Conference on Computer Vision (ICCV)}, 2021.

\bibitem[Rezende and Mohamed(2015)]{norm1}
Danilo~Jimenez Rezende and Shakir Mohamed.
\newblock Variational inference with normalizing flows.
\newblock In \emph{Proceedings of the 32nd International Conference on International Conference on Machine Learning - Volume 37}, page 1530–1538. JMLR.org, 2015.

\bibitem[Salzmann et~al.(2023)Salzmann, Chiang, Ryll, Sadigh, Parada, and Bewley]{pose_aut3}
Tim Salzmann, Hao-Tien~Lewis Chiang, Markus Ryll, Dorsa Sadigh, Carolina Parada, and Alex Bewley.
\newblock Robots that can see: Leveraging human pose for trajectory prediction.
\newblock \emph{IEEE Robotics and Automation Letters}, 8\penalty0 (11):\penalty0 7090--7097, 2023.

\bibitem[Sharma et~al.(2019)Sharma, Varigonda, Bindal, Sharma, and Jain]{Sharma_2019_ICCV}
Saurabh Sharma, Pavan~Teja Varigonda, Prashast Bindal, Abhishek Sharma, and Arjun Jain.
\newblock Monocular 3d human pose estimation by generation and ordinal ranking.
\newblock In \emph{The IEEE International Conference on Computer Vision (ICCV)}, 2019.

\bibitem[Shin et~al.(2023)Shin, Kim, Halilaj, and Black]{wham}
Soyong Shin, Juyong Kim, Eni Halilaj, and Michael~J. Black.
\newblock Wham: Reconstructing world-grounded humans with accurate 3d motion.
\newblock \emph{2024 IEEE/CVF Conference on Computer Vision and Pattern Recognition (CVPR)}, pages 2070--2080, 2023.

\bibitem[von Marcard et~al.(2018)von Marcard, Henschel, Black, Rosenhahn, and Pons-Moll]{3dpw}
Timo von Marcard, Roberto Henschel, Michael Black, Bodo Rosenhahn, and Gerard Pons-Moll.
\newblock Recovering accurate 3d human pose in the wild using imus and a moving camera.
\newblock In \emph{European Conference on Computer Vision (ECCV)}, 2018.

\bibitem[Von~Marcard et~al.(2018)Von~Marcard, Henschel, Black, Rosenhahn, and Pons-Moll]{von2018recovering}
Timo Von~Marcard, Roberto Henschel, Michael~J Black, Bodo Rosenhahn, and Gerard Pons-Moll.
\newblock Recovering accurate 3d human pose in the wild using imus and a moving camera.
\newblock In \emph{Proceedings of the European conference on computer vision (ECCV)}, pages 601--617, 2018.

\bibitem[Wehrbein et~al.(2021)Wehrbein, Rudolph, Rosenhahn, and Wandt]{WehRud2021}
Tom Wehrbein, Marco Rudolph, Bodo Rosenhahn, and Bastian Wandt.
\newblock Probabilistic monocular 3d human pose estimation with normalizing flows.
\newblock In \emph{International Conference on Computer Vision (ICCV)}, 2021.

\bibitem[Weng et~al.(2024)Weng, Bravo-S{\'a}nchez, and Yeung-Levy]{weng2024diffusion}
Zhenzhen Weng, Laura Bravo-S{\'a}nchez, and Serena Yeung-Levy.
\newblock Diffusion-hpc: Synthetic data generation for human mesh recovery in challenging domains.
\newblock In \emph{2024 International Conference on 3D Vision (3DV)}, pages 257--267. IEEE, 2024.

\bibitem[Yang et~al.(2023)Yang, Cai, Mei, Liu, Chen, Xiao, Wei, Qing, Wei, Dai, et~al.]{yang2023synbody}
Zhitao Yang, Zhongang Cai, Haiyi Mei, Shuai Liu, Zhaoxi Chen, Weiye Xiao, Yukun Wei, Zhongfei Qing, Chen Wei, Bo Dai, et~al.
\newblock Synbody: Synthetic dataset with layered human models for 3d human perception and modeling.
\newblock In \emph{Proceedings of the IEEE/CVF International Conference on Computer Vision}, pages 20282--20292, 2023.

\bibitem[Yu et~al.(2021)Yu, Salzmann, Fua, and Rhodin]{pcls}
Frank Yu, Mathieu Salzmann, Pascal Fua, and Helge Rhodin.
\newblock Pcls: Geometry-aware neural reconstruction of 3d pose with perspective crop layers.
\newblock In \emph{2021 IEEE/CVF Conference on Computer Vision and Pattern Recognition (CVPR)}, pages 9060--9069, 2021.

\bibitem[Yuan et~al.(2022)Yuan, Iqbal, Molchanov, Kitani, and Kautz]{glamr}
Ye Yuan, Umar Iqbal, Pavlo Molchanov, Kris Kitani, and Jan Kautz.
\newblock Glamr: Global occlusion-aware human mesh recovery with dynamic cameras.
\newblock In \emph{Proceedings of the IEEE/CVF Conference on Computer Vision and Pattern Recognition (CVPR)}, 2022.

\bibitem[Zanfir et~al.(2022)Zanfir, Zanfir, Gorban, Ji, Zhou, Anguelov, and Sminchisescu]{pose_aut1}
Andrei Zanfir, Mihai Zanfir, Alex Gorban, Jingwei Ji, Yin Zhou, Dragomir Anguelov, and Cristian Sminchisescu.
\newblock {HUM}3{DIL}: Semi-supervised multi-modal 3d humanpose estimation for autonomous driving.
\newblock In \emph{6th Annual Conference on Robot Learning}, 2022.

\bibitem[Zeng et~al.(2022)Zeng, Yang, Ju, Li, Wang, and Xu]{smoothnet}
Ailing Zeng, Lei Yang, Xuan Ju, Jiefeng Li, Jianyi Wang, and Qiang Xu.
\newblock Smoothnet: A plug-and-play network for refining human poses in videos.
\newblock In \emph{European Conference on Computer Vision}. Springer, 2022.

\bibitem[Zhang et~al.(2020)Zhang, Huang, and Wang]{3doh50k}
Tianshu Zhang, Buzhen Huang, and Yangang Wang.
\newblock Object-occluded human shape and pose estimation from a single color image.
\newblock In \emph{IEEE Conference on Computer Vision and Pattern Recognition, (CVPR)}, 2020.

\bibitem[Zhang et~al.(2023)Zhang, Ji, Kortylewski, Wang, Mei, and Yuille]{3dnbf}
Yi Zhang, Pengliang Ji, Adam Kortylewski, Angtian Wang, Jieru Mei, and Alan~L Yuille.
\newblock {3D-Aware Neural Body Fitting for Occlusion Robust 3D Human Pose Estimation}.
\newblock In \emph{The IEEE/CVF International Conference on Computer Vision}, 2023.

\bibitem[Zhou et~al.(2015)Zhou, Zhu, Leonardos, Derpanis, and Daniilidis]{Zhou2015SparsenessMD}
Xiaowei Zhou, Menglong Zhu, Spyridon Leonardos, Konstantinos~G. Derpanis, and Kostas Daniilidis.
\newblock Sparseness meets deepness: 3d human pose estimation from monocular video.
\newblock \emph{2016 IEEE Conference on Computer Vision and Pattern Recognition (CVPR)}, pages 4966--4975, 2015.

\bibitem[Zhu et~al.(2023)Zhu, Ma, Liu, Liu, Wu, and Wang]{motionbert2022}
Wentao Zhu, Xiaoxuan Ma, Zhaoyang Liu, Libin Liu, Wayne Wu, and Yizhou Wang.
\newblock Motionbert: A unified perspective on learning human motion representations.
\newblock In \emph{Proceedings of the IEEE/CVF International Conference on Computer Vision}, 2023.

\bibitem[Zwicker et~al.(2001)Zwicker, Pfister, Van~Baar, and Gross]{zwicker2001surface}
Matthias Zwicker, Hanspeter Pfister, Jeroen Van~Baar, and Markus Gross.
\newblock Surface splatting.
\newblock In \emph{Proceedings of the 28th annual conference on Computer graphics and interactive techniques}, pages 371--378, 2001.

\end{thebibliography}
